\pdfoutput=1

\documentclass[11pt]{article}

\usepackage{acl}

\usepackage{times}
\usepackage{latexsym}

\usepackage{graphics, amsmath}
\usepackage{todonotes}
\usepackage{hyperref}
\usepackage{xspace}
\usepackage{siunitx}
\usepackage{subcaption}
\usepackage{caption}
\usepackage{enumitem}
\usepackage{multirow}
\usepackage{booktabs}
\usepackage{xcolor}

\usepackage[T1]{fontenc}

\usepackage[utf8]{inputenc}

\usepackage{microtype}

\newif{\ifhidecomments}
\ifhidecomments
    \newcommand{\viv}[1]{}
    \newcommand{\alison}[1]{}
    \newcommand{\ke}[1]{}
    \newcommand{\joel}[1]{}
    \newcommand{\isabel}[1]{}
    \newcommand{\regina}[1]{}
    \newcommand{\alex}[1]{}
\else
    \newcommand{\viv}[1]{\textcolor{magenta}{[#1 ---\textsc{viv}]}}
    \newcommand{\alison}[1]{\textcolor{blue}{[#1 ---\textsc{alison}]}}
    \newcommand{\ke}[1]{\textcolor{red}{[#1 ---\textsc{ke}]}}
    \newcommand{\joel}[1]{\textcolor{orange}{[#1 ---\textsc{joel}]}}
    \newcommand{\isabel}[1]{\textcolor{purple}{[#1 ---\textsc{isabel}]}}
    \newcommand{\regina}[1]{\textcolor{green}{[#1 ---\textsc{regina}]}}
    \newcommand{\alex}[1]{\textcolor{cyan}{[#1 ---\textsc{alex}]}}
\fi

\newcommand{\para}[1]{\noindent{\bf #1}}

\newcommand{\figref}[1]{Fig.~\ref{#1}}
\newcommand{\secref}[1]{\S\ref{#1}}
\newcommand{\tabref}[1]{Table~\ref{#1}}

\newcommand{\condmanual}{\textit{Manual}\xspace}
\newcommand{\condai}{\textit{AI post-edit}\xspace}
\newcommand{\condhuman}{\textit{Human post-edit}\xspace}
\newcommand{\condref}{\textit{Human reference}\xspace}
\newcommand{\condgen}{\textit{AI-generated}\xspace}

\newcommand{\datareddit}{Reddit\xspace}
\newcommand{\datareddittifu}{Reddit-TIFU\xspace}
\newcommand{\dataxsum}{XSum\xspace}

\graphicspath{{figs/}}

\title{An Exploration of Post-Editing Effectiveness in Text Summarization}

\author{
{\bf Vivian Lai}\thanks{~~ Research work was done while authors were interning at Dataminr Inc.} ~$^1$, {\bf Alison Smith-Renner} $^2$, {\bf Ke Zhang} $^2$, {\bf Ruijia Cheng} \footnotemark[1] ~$^3$, \\
{\bf Wenjuan Zhang} $^2$, {\bf Joel Tetreault} $^2$, {\bf Alejandro Jaimes} $^2$ \\ 
$^1$ University of Colorado Boulder, \texttt{vivian.lai@colorado.edu} \\ 
$^3$ University of Washington, \texttt{rcheng6@uw.edu} \\ 
$^2$ Dataminr Inc., \texttt{\{arenner,kzhang,wzhang,jtetreault,ajaimes\}@dataminr.com} \\
}

\begin{document}
\maketitle

\begin{abstract}

%
Automatic summarization methods are efficient but can suffer from low quality.
In comparison, manual summarization is expensive but produces higher quality.
Can humans and AI \textit{collaborate} to improve summarization performance?
In similar text generation tasks (e.g., machine translation), human-AI collaboration in the form of ``post-editing'' AI-generated text reduces human workload and improves the quality of AI output. 
Therefore, we explored whether post-editing offers advantages in text summarization.
Specifically, we conducted an experiment with 72 participants, comparing post-editing provided summaries with manual summarization for summary quality, human efficiency, and user experience on formal (\dataxsum news) and informal (\datareddit posts) text.
%
%
%
%
This study sheds valuable insights on when post-editing is useful for text summarization: it helped in some cases (e.g., when participants lacked domain knowledge) but not in others (e.g., when provided summaries include inaccurate information). 
Participants' different editing strategies and needs for assistance offer implications for future human-AI summarization systems.  
\end{abstract}


\section{Introduction}
\label{sec:intro}
Text summaries provide short overviews of long documents or document collections, allowing readers to understand the content without the need to read full documents.
For example, news summaries outline key points so that readers do not have to read the entire article.   
For scientific papers, abstracts allow readers to easily understand the extent of the work and decide whether the paper is relevant to them.  
%
%
While these human-written summaries are typically high quality, a human's time and energy is limited and such tasks require heavy cognitive load~\citep{kirkland1991maximizing}.
%

Therefore, increasing research effort has explored machine models to generate summaries automatically~\cite{tas2007survey,nenkova2012survey, ELKASSAS2021113679}.
%
%
While recent advances in learning algorithms and data have resulted in models that can generate relatively high-quality summaries, human summarization is still the gold standard.
Further, training models require large, high-quality summarization datasets that are expensive to curate.

Taking advantage of the complementary strengths of humans and AI, can they \textit{collaborate} to improve summarization performance? 
%
In the area of machine translation, a common method of human-AI collaboration is human post-editing of AI-generated text, which increases human productivity and improves the quality of translation~\citep{koponen2016machine,vieira2019post}. 
However, in spite of its potential impact, studies of post-editing for summarization have been very limited, e.g. \citet{moramarco2021preliminary}, in the medical domain.
%

%
%
%

To bridge this gap, we performed a large-scale human subject experiment  (72 participants) investigating the utility of post-editing provided summaries in text summarization for informal (Reddit) and formal (news) datasets.
%
We expect tradeoffs in quality and efficiency (i.e., it might take longer to write better summaries), so we are interested in whether post-editing can actually improve efficiency or quality over manual methods, as well as the effects on users' experience.
This work is an important step 
toward understanding the benefits and drawbacks of post-editing as opposed to manual text summarization.

Our main contributions can be summarized as follows:
(1) we present the first large, human subjects experiment of post-editing for text summarization; (2) we show how post-editing impacts summary quality, efficiency, and user experience---where it is useful and where it is not; and (3) we create, and make public, two new datasets, each with 360 human-evaluated summaries for news and Reddit posts---either written manually or post-edited on provided summaries.

\section{Related work}
\label{sec:related}

\subsection{Automatic Text Summarization}
The field of automatic text summarization can be traced back to 1950s \citep{luhn1958automatic};
and since then much research has been devoted to developing algorithms, datasets, and evaluation metrics to developing summarization systems that can approach the quality of a human ~\cite{ELKASSAS2021113679}.
%
There are two primary automatic methods: (1) \textit{extractive}~\citep{dorr2003hedge,Nallapati2017SummaRuNNerAR}, where the model selects important sentences from the input document, and (2) \textit{abstractive}~\citep{rush2015neural,paulus2017deep}, where important parts of the input document are paraphrased to form new sentences.
While recent deep learning-based summarization methods have significantly advanced the quality of AI-generated summaries, they face some common issues, including hallucination, also known as contextual inconsistency~\citep{maynez2020faithfulness} and factual inconsistency~\citep{cao2018faithful,kryscinski2019evaluating}.
%
These critical issues limit the utility of automatic summarization if unaddressed; in fact, 
humans can be in the loop to manually fix identified mistakes, thus iteratively improving AI models~\citep{zhang2012active, gidiotis2021uncertainty}.

\subsection{Human Text Summarization}
Summaries written by humans often serve as gold standard references to train and evaluate AI models \cite{bhandari-etal-2020-evaluating}. One natural source of human summaries is shared and collected on the web, e.g., titles of news articles \cite{DBLP:conf/acl/SeeLM17, narayan2018don}, TL;DR of Reddit posts \cite{volske2017tl,kim2018abstractive}, talk transcripts of scientific papers \cite{lev-etal-2019-talksumm}, and government bill summaries \cite{kornilova-eidelman-2019-billsum}. 
Those are generated to serve specific goals and audiences, and often contribute datasets at scale to build AI models, although the data quality is not guaranteed
~\cite{kryscinski-etal-2019-neural,bommasani-cardie-2020-intrinsic}. 
Alternatively, human summaries can be annotated by dedicated professionals or crowd-workers for domain-specific documents \cite{jiang-etal-2018-effective}.
Annotators are often trained with guidance and summarization criteria, yet quality control~\citep{daniel2018quality}, due to subjectivity and inconsistency between annotators~\citep{tang2021investigating}, is a challenge. 
While the annotation process is costly and time-consuming,
human summarization, often by domain experts, yields higher quality compared to automatic methods \cite{zhang2020pegasus}. 
In this work, we turn to a common method of human-AI collaboration---human post-editing of AI-generated text---as an exploration for a viable solution.


\subsection{Post-Editing AI-Generated Text}
Post-editing is a common technique in machine translation, where translators edit the translations produced by automatic methods (opposed to completing the translations manually).
%
It has been shown to increase productivity and improve translation quality~\cite{Plitt2010APT,koponen2016machine,vieira2019post}, particularly when initial translations are good.  However, post-editing longer segments can require more cognitive effort to identify errors and plan corrections~\cite{koponen2012comparing}.

As summarization shares similarities to machine translation, post-editing is a promising paradigm, yet it is underexplored. 
%
One exception is \citet{moramarco2021preliminary}, who evaluated post-editing in the medical domain.
In a study with 3 physicians, participants took less time to post-edit other physician's written notes as compared to AI-generated notes, and post-editing any type of notes was faster than writing an entire note from scratch.
%
Participants' 
note-taking style differences also affected post-editing time.
For example, Doctor A wrote shorter notes and only edited AI-generated notes when there were substantial issues while Doctor B was more meticulous and edited the AI-generated notes extensively.
We build on prior work and explore post-editing for text summarization at a larger scale (72 participants) over two domains.
\section{Evaluating Post-Editing for Text Summarization}
\label{sec:method}
We explored
how providing summaries for post-editing affects (RQ1) final summary quality, (RQ2) efficiency, and (RQ3) user experience, compared to fully manual or fully automatic approaches for two domains: social media and news. 
Participants reviewed documents and summarized them, either without any assistance (manual) or provided with a human-written or AI-generated summary that they could edit (post-edit).
A distinct set of annotators then evaluated the quality of the summaries.
%
We included both human-written and AI-generated summaries in our study to explore post-editing for different summary types and qualities. 
\subsection{Data and Model}
We chose social media posts and news articles for our study as they could be understood by a general audience and are commonly experimented with automatic summarization literature. 
We also chose these datasets as they vary in writing formality, which might impact how humans understand and summarize text. 
Specifically, we used the \datareddittifu dataset~\citep{kim2018abstractive} (informal, Reddit ``Today I F'd Up'' posts) and the Extreme Summarization (\dataxsum) dataset~\citep{narayan2018don} (formal, British news articles). 
%
Each of these datasets includes  human-written ``reference'' summaries for the original documents: \datareddittifu uses the ``TL;DR'' written by the author of the post\footnote{Reddit users often self-summarize their posts with ``TL;DR:'' or ``too long; didn't read:'' statements.} while \dataxsum uses the introductory sentences---written by journalists---as the summaries (see \tabref{tb:example_summaries}).

For participants to summarize during our study,
we randomly selected 120 documents from the test sets (10 documents per participant, per condition),\footnote{\href{https://huggingface.co/datasets/reddit_tifu}{Reddit-TIFU}, \href{https://huggingface.co/datasets/xsum}{Xsum}} with length between the 25th and 75th percentile to balance task difficulty and time.
The average length of the \datareddit posts is 243.8 words, and the average length of the \dataxsum articles is 223.3 words (see Appendix \ref{app:doc_len_dist} for length distribution).

We used the Pegasus model~\cite{zhang2020pegasus} to generate summaries for the two datasets. 
Pegasus is a masked language model pre-trained with a novel self-supervised objective, gap-sentences generation, and fine-tuned on downstream abstractive summarization tasks. 
The model achieved state-of-the-art performance on multiple datasets, including \dataxsum and \datareddittifu. 
We directly applied the off-the-shelf Pegasus models downloaded from HuggingFace, with one already finetuned on \dataxsum \footnote{\href{https://huggingface.co/google/pegasus-xsum}{huggingface-pegasus-xsum}} and the other on \datareddittifu \footnote{\href{https://huggingface.co/google/pegasus-reddit_tifu}{huggingface-pegasus-reddit\_tifu}}.

We did not introduce summaries from any other models besides Pegasus, as the goal of this paper was not to compare models but to understand how human post-editing of provided summaries compares to manual and automatic methods. 
And, while Pegasus is currently high-performing compared to other, weaker models, the summaries we provided for post-editing in our study were of varied quality, particularly between datasets (see \secref{sec:results:quality}).
This gave us an opportunity to explore how summary (or assistance) quality might affect human post-editing.



\subsection{Study Design}
This study consists of two phases: (1) summary collection and (2) human evaluation of the collected summaries.
For summary collection, we used a between-subjects experimental design, with three conditions: (1) \condmanual, where participants wrote summaries without any assistance; (2) \condai, where participants post-edited AI-generated summaries; and (3) \condhuman, where participants post-edited human-written summaries.
Participants summarized either informal \datareddit posts or formal \dataxsum news articles.
%
For the \condhuman condition, participants were provided the human written ``reference'' summaries from each of the datasets. 
In the following, we describe the participants and procedure for the summary collection phase, followed by details of the evaluation phase.

\begin{table}[h!]
    \small
    \centering
    \begin{tabular}{p{0.1\textwidth}p{0.15\textwidth}p{0.15\textwidth}}
    \toprule
     & Condition & WPM (M, $\sigma$)\\ \midrule
    \datareddit & Manual & 334, 113 \\
     & AI post-edit & 374, 175 \\
     & Human post-edit & 363, 104 \\ \midrule
    \dataxsum & Manual & 327, 112 \\
     & AI post-edit & 360, 241 \\
     & Human post-edit & 396, 178 \\
    \bottomrule
    \end{tabular}
    \caption{Average reading speed per condition per dataset. WPM represents words per minute. 
    Strategic assignment to conditions ensured no significant differences between reading speed.
    }
    \label{tb:reading_speed}
\end{table}

\subsection{Summary Collection Participants}
We recruited 72 participants (45 female, 22 male, 3 non-binary, 2 preferred not to disclose) from  Upwork.\footnote{https://www.upwork.com; we used Upwork---as opposed to other crowdsourcing platforms---to recruit experienced participants and ensure higher quality summaries.}
They were on average 32 years old ($\sigma$=12) and were required to be native or bilingual English speakers, have at least a 90\% job success score, and possess expertise in writing, journalism, or communication.
%
To ensure participants had some familiarity with the summarization domains, they described their experience reading or posting on Reddit and knowledge of British news.
Specifically, participants rated the extent of the respective knowledge based on a 7-point Likert scale, and were selected if they responded at a rating of 4 or above.
Finally, participants reported their reading (in words per minute, WPM) and comprehension scores,\footnote{https://swiftread.com} which we used to (1) eliminate those with comprehension less than 75\% and (2) account for reading speed when assigning conditions.
To account for differences in participants' reading speed that could affect our results, we assigned participants into conditions, ensuring a similar average reading speed across conditions (\tabref{tb:reading_speed}).
%
%
The average reading speed for participants in our study was 358 WPM ($\sigma=158$). 

\subsection{Summary Collection Procedure}
\label{sec:procedure}
Based on pilot studies (see Appendix \ref{app:pilot}), we anticipated the summary collection task sessions would take an hour and we paid participants \$20.
Each participant took on average 33.9 minutes ($\sigma$=15.0) to perform the summarization task.\footnote{Participants spent additional time completing the post-task survey.}
Of the 72 participants, an equal number (12) were randomly assigned to each dataset (\datareddit or \dataxsum) and summarization condition (\condmanual, \condhuman, \condai), ensuring a similar average reading time for each condition (see \tabref{tb:reading_speed}).
During the study, participants completed three phases: (1) instructions, tutorial and practice, (2) summarization task, and (3) post-task survey.

\begin{figure}[t!]
    \centering
    \begin{subfigure}[t]{0.5\textwidth}
        \centering
        \includegraphics[width=\textwidth]{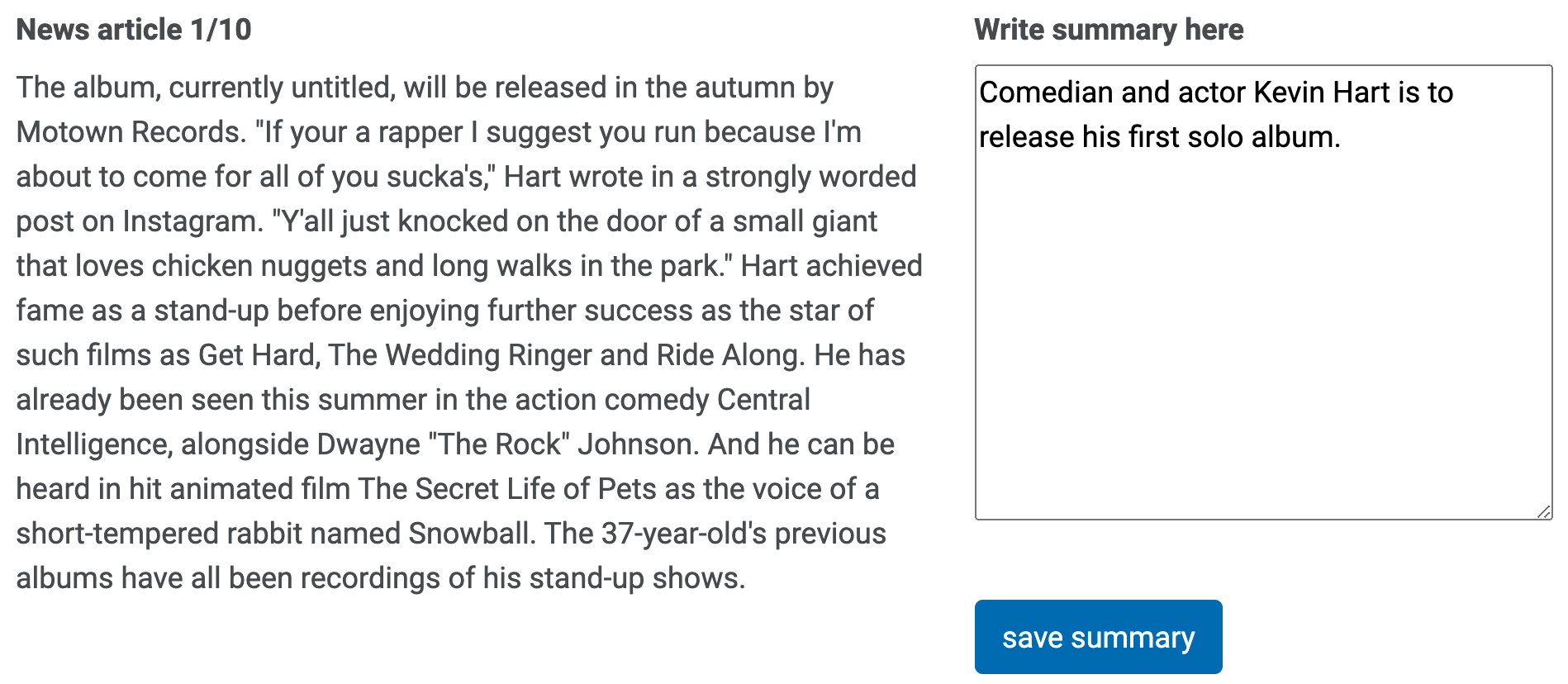}
    \end{subfigure}
    \caption{Sample task interface for the \condai condition for \dataxsum, showing the provided, AI-generated summary in the text box.}
    \label{fig:upwork_task_interface}
\end{figure}

Participants first reviewed task instructions, the criteria for writing a good summary (from \citet{stiennon2020learning}), and examples of good and bad summaries with explanations. 
For consistency, we used the same criteria when asking annotators (a distinct set of human evaluators) to evaluate the summary quality. 
%
Participants then reviewed 10 documents (either all \datareddit posts or all \dataxsum news articles, depending on their assignment) and summarized each, either manually (\condmanual) or post-editing a provided human summary (\condhuman) or AI-generated summary (\condai).
\figref{fig:upwork_task_interface} shows an example of the task interface for the \condai condition and \dataxsum.
Participants were not made aware of the source of their provided summaries--whether human or AI.
%
Per condition, each participant summarized a unique set of 10 documents.
Participants had access to the summarization criteria as guidance while summarizing.
After completing each summary, participants rated the difficulty for summarizing the original document.
Finally, after finishing all 10 summaries, participants completed a survey about their experience.
See \ref{app:phase1_procedure} for details on the procedure and tutorial examples.

\subsection{Human Evaluation of Summary Quality}
\label{sec:human-eval}
%
To evaluate the quality of the summaries written during our study, we recruited a distinct set of annotators from Amazon Mechanical Turk.
%
To ensure quality ratings, we only employed turkers who satisfied the following criteria: (1) completed 5000 HITs; (2) 97\% HIT approval rate; (3) reside in the United States, Australia, and United Kingdom.
%
Annotators underwent tutorials and multiple attention-check questions before performing the task (see \ref{app:evaluate_summary}).
We also eliminated annotators with validation procedures (see \ref{app:quality_control}).
Annotators were paid \$1.50 per HIT (see \ref{app:evaluate_summary}) and were allowed to perform multiple HITs, assuming they would improve at the evaluation task over time. 
Each annotator performed 9.4 HITs on average.

The annotators evaluated six different summaries for each original document (\dataxsum article or \datareddit post) from our study: 
(1) the \condmanual summary written without any assistance,
(2) the summary written in the \condai condition, given (3) the \textit{AI-generated} summary from the Pegasus model, 
(4) the summary written in the \condhuman condition,
given (5) the \textit{human reference} from the dataset,
and, finally, (6) a \textit{random} summary generated by randomly selecting two sentences from the opposite dataset.
See Table \ref{tb:example_summaries} in \ref{app:example_summary_type} for examples of each summary type.
The random summary was used as selection criteria to identify annotators who were not paying attention during the task. 
Following \citet{stiennon2020learning}, annotators evaluated each summary on four axes: \textit{coherence}, \textit{accuracy}, \textit{coverage}, and \textit{overall}. 
Each summary was evaluated by five annotators; we removed outliers then averaged the remaining annotators ratings to determine the final rating for each summary.
%
%
Refer to \ref{app:eval} for additional details on the human summary evaluation procedure, annotators, and quality control.


\subsection{Measures}
\label{para:measures}
We report on {\it summary quality}, {\it efficiency}, {\it user experience} for the summarization task. 
Summary quality is measured using human ratings and 
%
efficiency is measured by the amount of time to read and write a summary for each document.
We additionally measured user workload as \textit{edit distance} or the difference between the provided summary and the final summary.

We also report on three subjective user experience measures collected using 7 point rating scales (from strongly disagree to strongly agree) either on the task-level (at the end of the task) or instance-level (after each summary): \textit{task difficulty}, ``I found it difficult to summarize the article well.'' (instance-level); 
\textit{frustration}, or ``performing the summarization tasks was frustrating.'' (task-level); and \textit{assistance utility}, or ``the provided summaries were not useful to me when I was performing the summarization tasks'' (task-level).\footnote{Only participants in the post-editing conditions responded about the utility of the provided summaries.}
All task-level measures were paired with follow up ``why did you feel this way'' open-ended questions. 
%
%

\subsection{Data and Analysis}
72 participants wrote 720 summaries (manually or post-editing provided summaries) and 113 annotators evaluated those summaries, resulting in 6360 summary quality ratings; after removing outliers (see \ref{app:quality_control}). 
Averaging resulted in one final quality rating (on four axes) for each summary.
%
We make this dataset of summaries and their ratings public to promote future research.\footnote{https://github.com/vivlai/post-editing-effectiveness-summarization}

To find out if any statistical differences exist between the means of the conditions, we used one-way ANOVA for each objective and subjective (rating scale) measure.
Using post-hoc Tukey's HSD, we also performed pairwise comparison to determine which two conditions are significantly different.
%

We qualitatively coded the open-ended responses related to the subjective measures of frustration, task difficulty, and assistance utility, as well as responses on likes, dislikes, and desired improvements.
One researcher read the data to identify emergent codes, followed by a discussion period to merge and update the themes in the codebook. 
Two researchers then independently coded all the open-ended responses respectively, achieving a high average inter-rater reliability (Cohen’s $\kappa >= .85$) for each of the open-ended responses.
See \ref{app:thematic_coding} for details on thematic coding.
We refer to participants as P1-P72.

\section{Results}
\label{sec:results}
We report on the impact of post-editing on summary quality, efficiency, and user experience.

\subsection{Summary Quality}
\label{sec:results:quality}
\begin{figure}[t]
    \centering
    \begin{subfigure}[t]{0.5\textwidth}
        \centering
        \includegraphics[width=\textwidth]{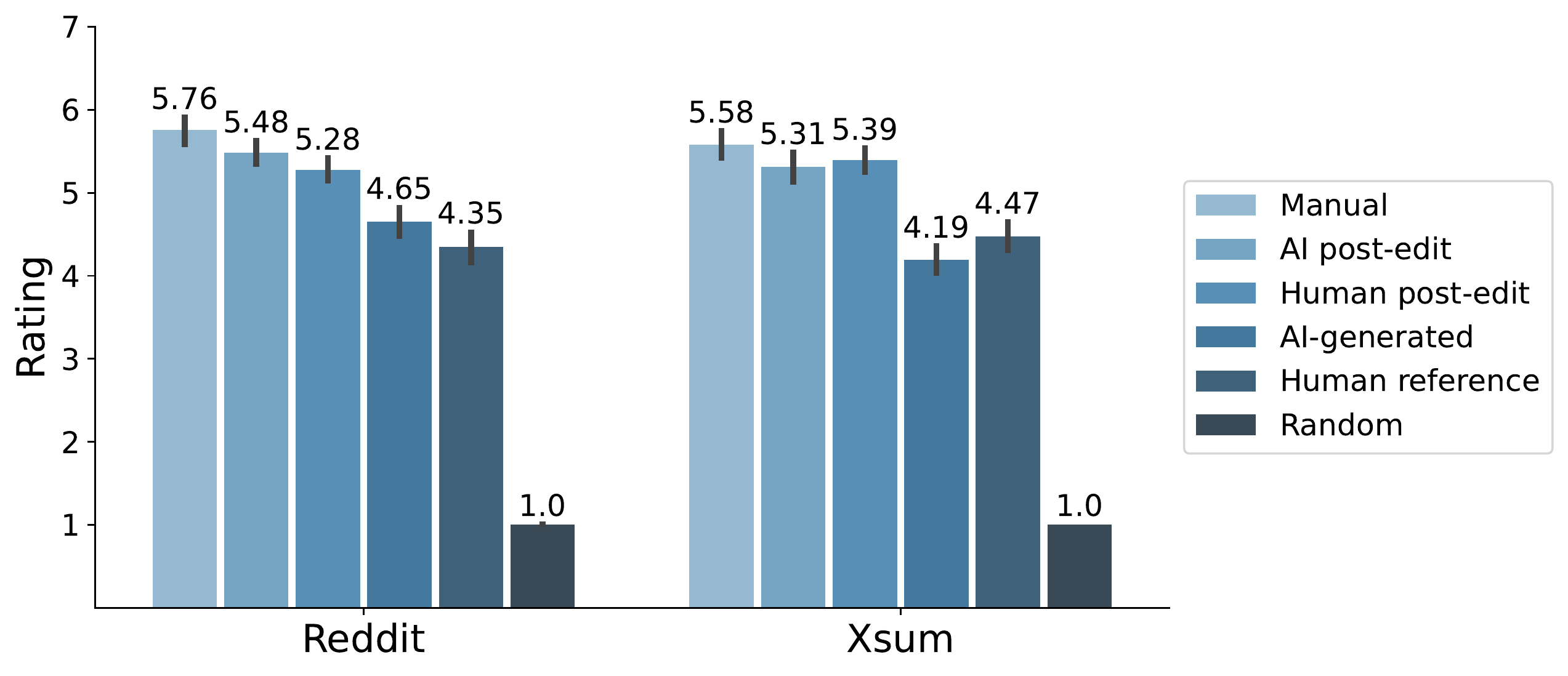}
    \end{subfigure}
    \caption{Average \textit{overall} quality ratings 
    for the summaries by type and dataset. For \datareddit, the human reference was the worst (aside from the Random summary). For \dataxsum, the AI-generated summary was the worst.}
    \label{fig:quality_overall_coherence}
\end{figure}

%
%
We discuss quality ratings for the summaries for each dataset (\figref{fig:quality_overall_coherence}).
For simplicity, we report only on \textit{overall} quality ratings from the human evaluation (see Appendix~\ref{app:acc_cov_ratings} for other axes).
%
%


\para{For \datareddit, post-editing improved the quality of the provided summaries but manual summaries were the best.}
\datareddit summaries produced by participants in the \condmanual condition 
were rated highest \textit{overall} quality; the provided summaries, \condgen 
followed by the \condref 
were the lowest quality (\figref{fig:quality_overall_coherence}).
Interestingly, our evaluation finds that the AI-generated summaries are significantly higher quality than the human references ($p=.02$). This is different from \citet{zhang2020pegasus},
for which the same Pegasus model achieved comparable performance to human references (but not better).

Comparing summarization conditions, we find significant differences for final summary quality ($p<.01, F=9.3$): \condmanual summaries outperformed summaries produced by participants in both the \condai ($p=.03$) and \condhuman ($p<.01$) conditions.
%
%
Finally, summaries resulting from \condai and \condhuman were significantly better than the provided summaries for those conditions, \condref ($p<.01$) and \condgen ($p<.01$), meaning participants improved the quality of the summaries they were given.

\para{For \dataxsum, post-editing improved the quality of the provided summaries and was just as good as manual.}
\dataxsum summaries produced by participants in the \condmanual condition were rated the highest and the provided summaries (\condref followed by \condgen) were the lowest (\figref{fig:quality_overall_coherence}).
However, for \dataxsum, summary quality was not significantly impacted by \textit{AI assistance} ($p=.08, F=2.5$), meaning there was no significant difference in quality between the \condmanual, \condai, or \condhuman summaries.

Similar to \datareddit, \condai and \condhuman summaries were significantly better than the provided summaries for those conditions ($p<.01$). 
But, opposite of \datareddit, the \dataxsum \condref summaries were significantly better than the \condgen summaries ($p=.02$).

\subsection{Efficiency}

\begin{figure}[t]
    \centering
    \begin{subfigure}[t]{0.5\textwidth}
        \centering
        \includegraphics[width=\textwidth]{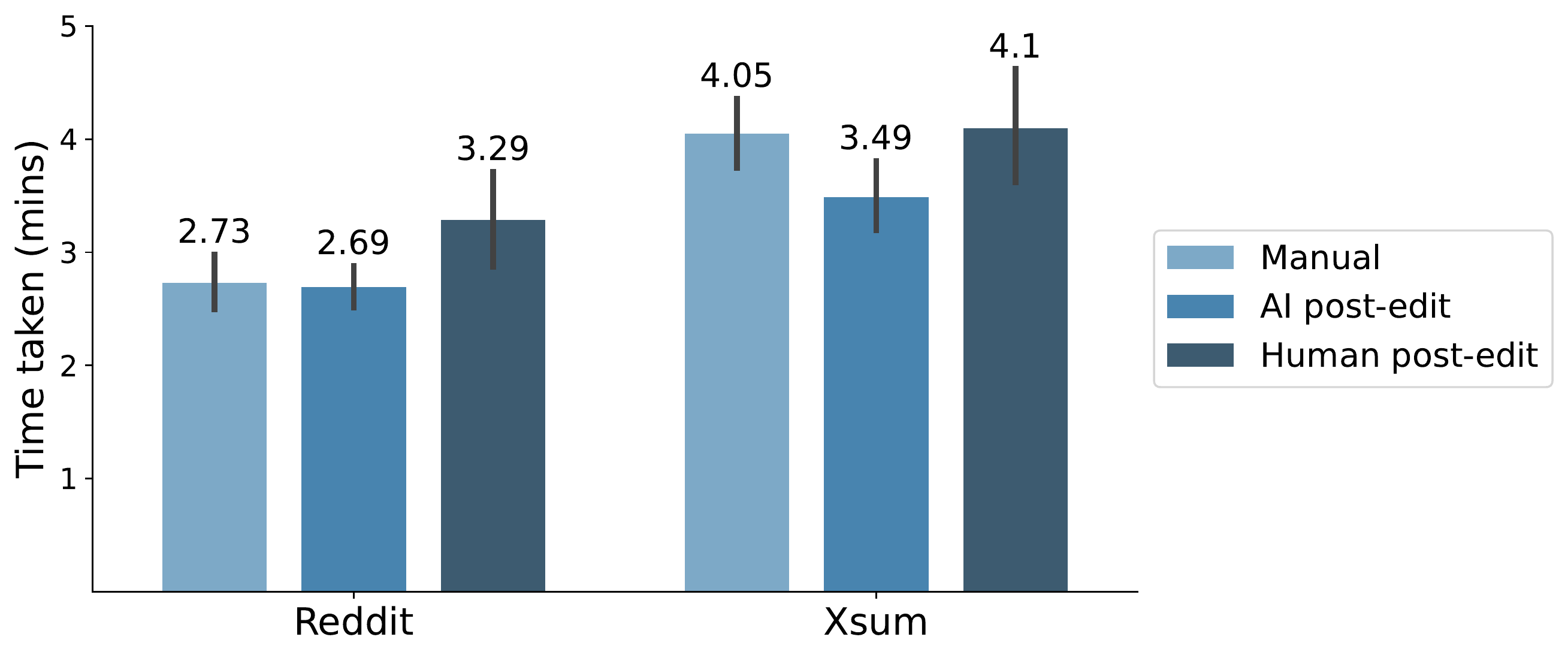}
    \end{subfigure}
    \centering
    \caption{Comparison between conditions for average time to summarize (per document) for \datareddit and \dataxsum. In general, participants in \dataxsum took longer to complete the task, likely due to unfamiliarity with the domain.
    }
    \label{fig:efficiency_comparison}
\end{figure}

\para{Post-editing human references slowed \datareddit summarization but post-editing AI-generated summaries was faster for \dataxsum.}
Summarization conditions significantly differed in efficiency (\figref{fig:efficiency_comparison}) for both \datareddit ($p=.04, F=3.2$) and \dataxsum ($p=.02, F=3.8$).
For \datareddit, it took significantly longer to post-edit human references (\condhuman) than manually (\condmanual, $p=.03$) or given AI-generated summaries (\condai, $p=.02$). We discuss possible reasons for this (e.g., problematic Reddit references) in \secref{sec:reddit-difficulty}.

For \dataxsum, post-editing AI-generated summaries (\condai) was faster than \condmanual ($p=.11$) or given human references (\condhuman, $p=.08$).
%
However, no pairwise comparisons were significant after correction.\footnote{Based on the trend, we would expect to see a significant result with more statistical power (more participants).}

\para{Provided summary quality did not impact the number of edits.}
Anticipating that participants might have needed to make more edits to improve on worse summaries, we compared the edit distance to provided summary quality (overall) using Spearman correlation.
However, for neither \datareddit nor \dataxsum, did summary quality have a strong relationship to edit distance.
In some cases participants made many edits to both good and bad summaries, whereas in others, they made very few edits regardless of quality (see \ref{app:edit_distance} for correlation plots), due in part to participants' diverse editing styles, where some desired to make changes regardless of the provided summary quality; \citet{moramarco2021preliminary} made similar observations.
%
For example, P10 (\datareddit, \condai) ``did not use [the provided summaries] at all'' and P55 (\datareddit, \condhuman) edited all the summaries to match their preferred writing style, stating ``I found the casual writing style confusing [...] I just did it my way.'' 

\begin{figure}[ht!]
    \centering
    \begin{subfigure}[t]{0.5\textwidth}
        \centering
        \includegraphics[width=\textwidth]{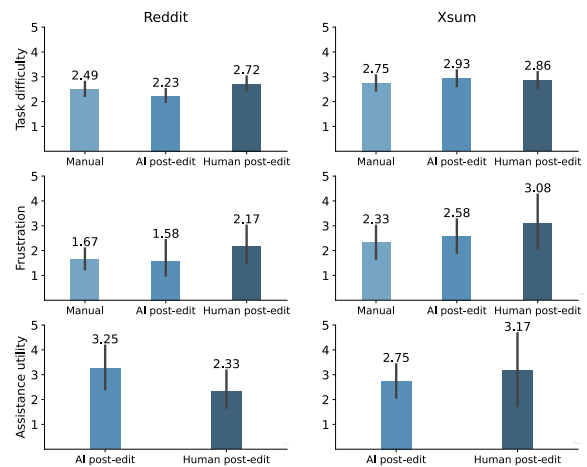}
    \end{subfigure}
    \centering
    \caption{User experience plots for task difficulty, ``I found it difficult to summarize the article well'', frustration, ``Performing the summarization tasks was frustrating'', and assistance utility, ``The provided summaries were not useful to me when I was performing the summarization tasks'' for \datareddit (Left) and \dataxsum (Right). Responses were collected using 7 point rating scales.}
    \label{fig:user_experience}
\end{figure}

\subsection{User Experience}
\label{sec:user_experience}
We measured user experience with \textit{task difficulty}, \textit{frustration}, and \textit{assistance utility} 
(\figref{fig:user_experience}).
We also surfaced insights about participants' experiences from the qualitative analysis. 


\para{Participants found it harder to post-edit \datareddit references.}
\label{sec:reddit-difficulty}
%
%
Summarization conditions significantly differed for \textit{task difficulty} for \datareddit ($p=.04, F=3.2$), but not for \dataxsum ($p=.72, F=.3$).
Specifically, summarizing when provided a \datareddit human reference (\condhuman) was perceived significantly more difficult than summarizing when provided an AI-generated summary (\condai, $p=.04$).
Other pairwise comparisons were not significant.
Recall that post-editing human references also took longer than other conditions for Reddit; this difficultly might be due to the fact that \datareddit human reference data consisted of poorly written TL;DRs, many of which add extra details not found in the original posts. As participants like P63 (\datareddit, \condhuman) and P58 (\datareddit, \condhuman) commented, some provided Reddit summaries were ``really bad'' or ``off a bit.''
%

%

%
%
%

\para{Participants were mixed on whether the provided summaries were useful.}
\label{sec:results:ux:utility}
%
%
\textit{Assistance utility} did not significantly differ between summarization condition for \datareddit ($p=.12, F=2.4$) or \dataxsum ($p=.63, F=.2$), due in part to the high variability in participants' responses.
However, participants provided mixed responses on the utility of the summaries: while many thought they were helpful ``starting points'' in their summarization process, others found they sometimes missed important points or contained information that was unneeded, incorrect, or incongruous with the original article. 
%

Participants, therefore, used the provided summaries in different ways. Some, like P59 (\datareddit, \condhuman) used the summary as a starting point or guideline and made edits on top of it, ``the provided summaries did the job pretty well, I just added some details.''
%
%
Others ignored the provided summary. As P61 (\datareddit, \condhuman) said, ``it might have been easier to do \textit{blind} summaries rather than having the provided examples.'' 
Some participants, like P31 (\dataxsum, \condhuman), chose to ``read the passage, write my summary, and then look at the given summary.'' 

Participants' had other concerns about post-editing the provided summaries during the task.
Some thought it took high cognitive load to make edits and summarize at the same time, ``it clouded my memory of what information the passage had actually provided'' (P31 \dataxsum, \condhuman).  
Others struggled with originality. While they perceived it ``a bit like cheating'' (P44 \datareddit, \condai) 
to use the provided summary instead of writing their own, many also ``found it difficult to provide a better summary than what was provided'' (P32 \dataxsum, \condai).
%
Finally, some participants noticed that they had the tendency to over-rely on the provided summary. For example, P6 (\dataxsum, \condhuman) found that they were distracted from their own thinking and unlikely to challenge the provided summary, ``the provided summaries deterred me from writing my own and gleaning my own major points from the articles. Instead, I would defer to the information given in the provided summaries and edit a few things, but not add anything major.''
\para{Comprehension of the original text can impact summarization.}
Participants found it challenging to summarize documents that lacked context, were overly detailed, or had poor quality. 
While we intentionally recruited participants with some expertise for the two data types (Reddit posts and British news), many were hindered by a lack of background knowledge, especially for the British news articles (\dataxsum). 
%
For example, P18 (\dataxsum, \condmanual) said it was difficult to summarize particular articles about ``the British government'' because ``it's not something I am familiar with so it was hard to determine what information to include in the summary.'' 
Similarly, P15 (\dataxsum, \condmanual) mentioned difficultly summarizing articles about Cricket, which contained ``sport-specific jargon or proper nouns that I was wholly unfamiliar with.'' 
In fact, 50\% of participants summarizing \dataxsum described a \textit{lack of contextual knowledge} compared to only 8\% of participants summarizing \datareddit posts.
%
Post-editing can help in this case by providing a useful starting point so that users do not need to fully understand the document and write manually.

Participants also found it difficult to decide what was important from overly detailed original documents, particularly when summarizing manually. For example, P17 (\dataxsum, \condmanual) stated, ``some articles gave so many details and it took a while to decide which were important to keep in a summary.''
Finally, participants were hindered by the poor quality of the original documents. 
For example, P60 (\datareddit, \condhuman) viewed the ``lack of capitalization and proper punctuation that is common with Reddit posts'' as the greatest frustration in the summarization process.
Participants found it challenging to match the tone and style of these documents in their summaries. 
For example, in P46's (\datareddit, \condai) words, it is ``hard to match in the winding, anecdotal writing style often found on Reddit.'' 
%
%


\subsection{Summary}
Post-editing yielded better quality summaries than the automatic methods. 
However, compared to manual summarization, the results were mixed.
For formal news articles, post-editing lead to similar quality summaries with improved efficiency, helping when participants lacked domain knowledge.
However, post-editing produced worse summaries, more slowly for Reddit posts, likely due to the informal writing style and sometimes inaccurate TL;DR references provided in that case.
%
%
%
%
%
We did not find a correlation between edit distance and provided summary quality, instead, some participants tended to make more edits---due to style or writing preferences---while others made fewer edits, instead of relying on the provided summaries---regardless of the quality of the summary they were editing. 
%
%




\section{Discussion}





\label{sec:discussion}

This work is the first large-scale study of post-editing for text summarization, providing valuable insights on the benefits and drawbacks.
We discuss these, as well as outline future research directions and design recommendations for post-editing summarization systems. Finally, we discuss the limitations of our experiments. 

%

\para{Post-editing was useful when domain context is needed.}
As our participants were not well-versed in the British news content of \dataxsum, the provided summary was ``useful'' as a starting point (as described in qualitative responses), so that they did not have to write from scratch.
This is similar to machine translation literature, which suggests that monolingual editors, despite lacking the knowledge of the other language, can still effectively improve the quality of translation via post-editing~\citep{koehn2010enabling}.
%
Beyond post-editing, systems could provide additional support when users lack domain knowledge or context, such as inline web searching to learn about unknown terms or phrases (e.g., the rules of Cricket).  

\para{Post-editing was less useful when the provided summaries were low quality.}
Low quality, particularly inaccurate or incoherent, provided summaries can be confusing and hard to edit, making them less ``useful'' as summarization assistance.
Ideally, such summaries are not provided for post-editing, as manual summarization would be better in those cases.
Future systems should explore techniques for determining \textit{whether} to provide a summary or not, based on desired summary qualities.
Finally, systems might provide \textit{transparency}, e.g., highlighting the important details in the original text~\citep{lai2019human}.
Then users can then decide for themselves whether or not those details are important and the summary should be trusted or ignored.

\para{Post-editing can lead to over-reliance and stifle creativity.}
Humans have a tendency to over-rely on AI systems~\cite{bussone2015role,buccinca2021trust}: 
prior work on text generation found users consider the provided text as an ``authority'' and thus feel apprehensive to make significant edits~\citep{bhat2021people}.
In our experiment, some participants reported a similar tendency to over-rely on the provided summary or were distracted from writing their own version of the summary; some even developed their own combative strategies: writing their summaries manually first and then referring to the provided assistance.
Therefore, future systems might allow different workflows, where summaries are shown before or after manual summarization (or not at all).
%

\para{Post-editing systems should cater to varied users’ preferences and needs.}
Users have varied summarization strategies and needs for assistance as a result of personal preferences and their experience with the domain. 
In our study, some preferred more control over the final summary, making lots of edits, while others made fewer edits, and a few did not use the provided summaries at all.
These differences in writing style (and their effects on post-editing in text summarization) are also noted in prior work~\citep{moramarco2021preliminary}.
Therefore, systems should give users control over the assistance they receive and alternative workflows, based on their preferences and needs. 

Users' needs also vary by their target audience; users might desire summaries that are longer or shorter, or more formal or informal based on who they expect will read them.
Post-editing can help, allowing users to tailor summaries to different audiences with the same underlying content.
Future post-editing systems might provide multiple summary options, with diverse content and/or style, from which users can choose.

\para{Limitations.}
We note possible limitations of our results due to the length and nature of the task: participants only interacted with the summarization system for a short time (less than an hour) and for a task, for which they lacked ownership; also, including more participants would have given more statistical power for comparing conditions.
Future work should perform experiments with more realistic and longer-term summarization engagements. 
Regarding our datasets, we chose two to differentiate between formal writing (i.e., news articles) and informal writing (i.e., social media posts).
However, we did not experiment with more societally critical summarization tasks, such as medical or legal documents. 
While post-editing was useful when more domain context was needed, it is unclear how our findings would generalize to more high-risk scenarios.

\section{Conclusion}

\label{sec:conclusion}
To take advantage of the complimentary strengths of AI---which can produce summaries quickly---and humans---which can write summaries well---we explored how human-AI collaboration (i.e. post-editing) impacts summary quality, human efficiency, and user experience for text summarization. 
%
Through the first large-scale study on post-editing for text summarization, we provide valuable insights on the benefits and drawbacks: 
compared to summarizing manually, post-editing was helpful for formal news articles, where participants lacked domain knowledge, while post-editing was less helpful for informal social media posts, for which the reference TL;DR summaries sometimes included inaccurate information.
We also observed differences in participants editing strategies and needs as well as concerns of over reliance, all of which deserve future exploration. 
%
%
%
%
We hope this initial exploration provides a starting point for future research on post-editing in text summarization.

\clearpage
\bibliographystyle{ACM-Reference-Format}
\bibliography{refs}

\clearpage
\appendix
\section{Appendix}
\label{sec:appendix}
Our study explored how providing summaries for post-editing affects summary quality, efficiency, and user experience compared to fully manual or fully automatic approaches. 
The study involved two phases: (1) summary collection (Appendix~\ref{app:study}) and (2) human evaluation of the collected summaries (Appendix~\ref{app:eval}). 
We also report on additional results (Appendix~\ref{app:results}) 

\subsection{Summary Collection}
\label{app:study}
We collected summaries through a summarization task, where participants first reviewed documents (from either \datareddit or \dataxsum) and summarized them, either without any assistance or provided with a human-written or model-generated summary they could edit. 

In the following, we describe details on the documents included in our study, how we piloted the task and interface, and more information about the study procedure.

\subsubsection{Document Length Distribution}
\label{app:doc_len_dist}

\begin{figure}[h!]
    \centering
    \begin{subfigure}{0.5\textwidth}
        \centering
        \includegraphics[width=\textwidth]{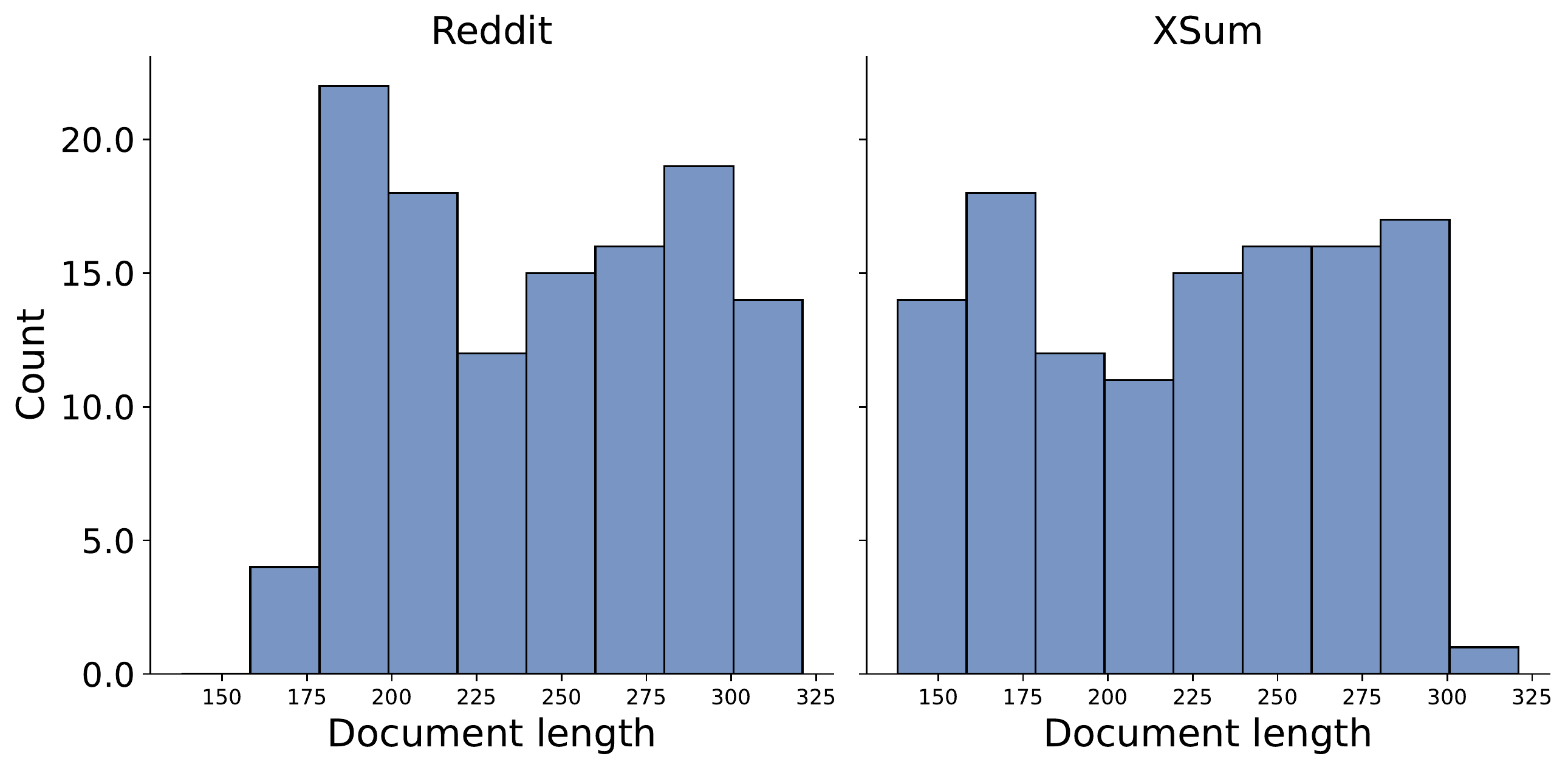}
    \end{subfigure}
    \caption{This figure shows the document length of distribution of both datasets. The average length of the \datareddit posts is 243.8 words, and the average length of the \dataxsum articles is 222.3 words.}
    \label{fig:doc_len_dist}
\end{figure}

Participants summarized 120 documents from the test sets (10 documents per participant, per condition),\footnote{\href{https://huggingface.co/datasets/reddit_tifu}{Reddit-TIFU}, \href{https://huggingface.co/datasets/xsum}{Xsum}} with length between the 25th and 75th percentile to balance task difficulty and time. Figure~\ref{fig:doc_len_dist} gives the length distribution for both datasets.

\subsubsection{Piloting the Summarization Task and Interface}
\label{app:pilot}
We performed two pilot study sessions (with researchers and Upwork pilot participants) for feedback on the web application, procedure, and to estimate session duration.
%
The first was conducted among five researchers from our lab and the second was conducted with 12 representative 
users from Upwork (see \secref{sec:method}), which were later included in the main study.

\begin{table}[h!]
    \small
    \centering
    \begin{tabular}{p{0.07\textwidth}p{0.38\textwidth}}
    \toprule
    Criteria & Explanation \\ 
    \midrule
    
    Essence & The summary is a good representation of the post. \\ 
    
    Clarity & The summary is reader-friendly. It expresses ideas clearly. \\ 
    
    Accuracy & The summary contains the same information as the longer post \\ 
    
    Purpose & The summary serves the same purpose as the original post. \\ 
    
    
    Style & The summary is written in the same style as the original post. \\
    \bottomrule
    \end{tabular}
    \caption{We showed this set of summary criteria to the participants in both tutorial and actual task.}
    \label{tb:summary_criteria}
\end{table}

\begin{figure}[h!]
    \centering
    \begin{subfigure}{0.5\textwidth}
        \centering
        \includegraphics[width=\textwidth]{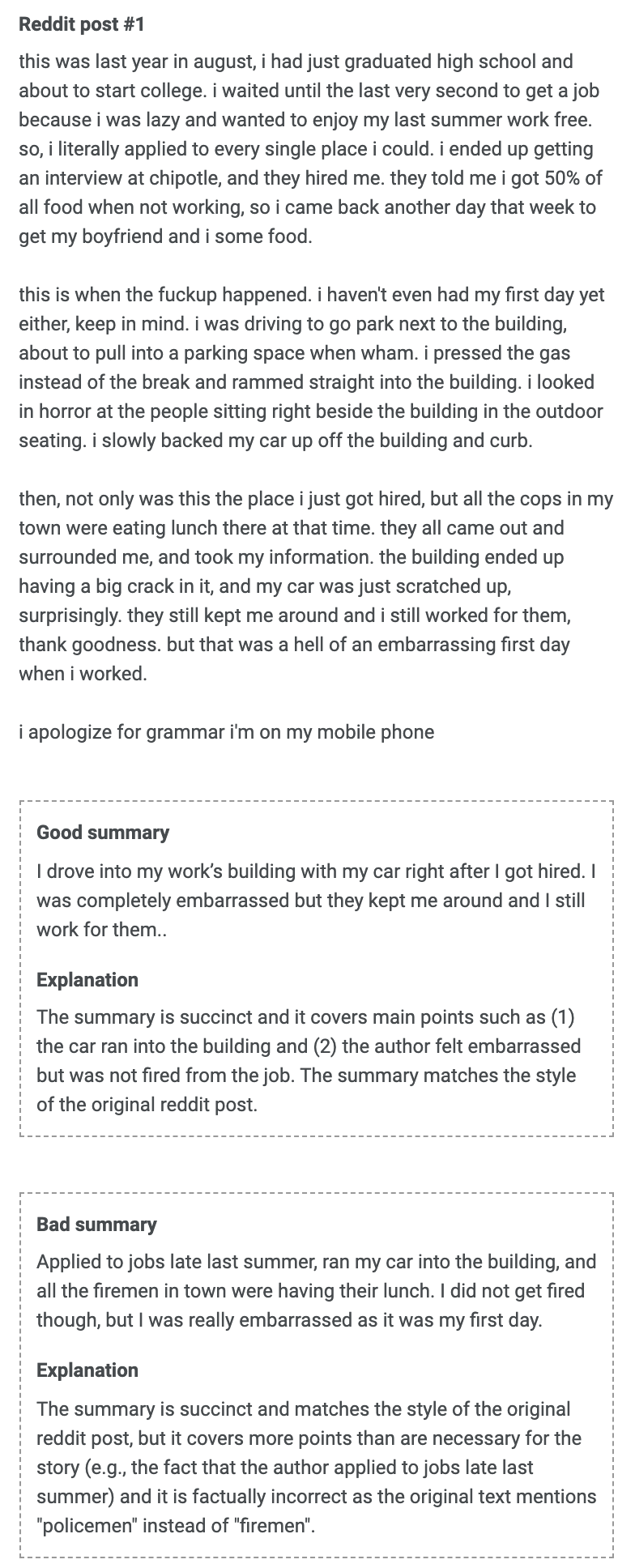}
    \end{subfigure}
    \caption{\datareddit tutorial first example with good and bad explanations.}
    \label{fig:tutorial_example1}
\end{figure}

\begin{figure}[h!]
    \centering
    \begin{subfigure}{0.5\textwidth}
        \centering
        \includegraphics[width=\textwidth]{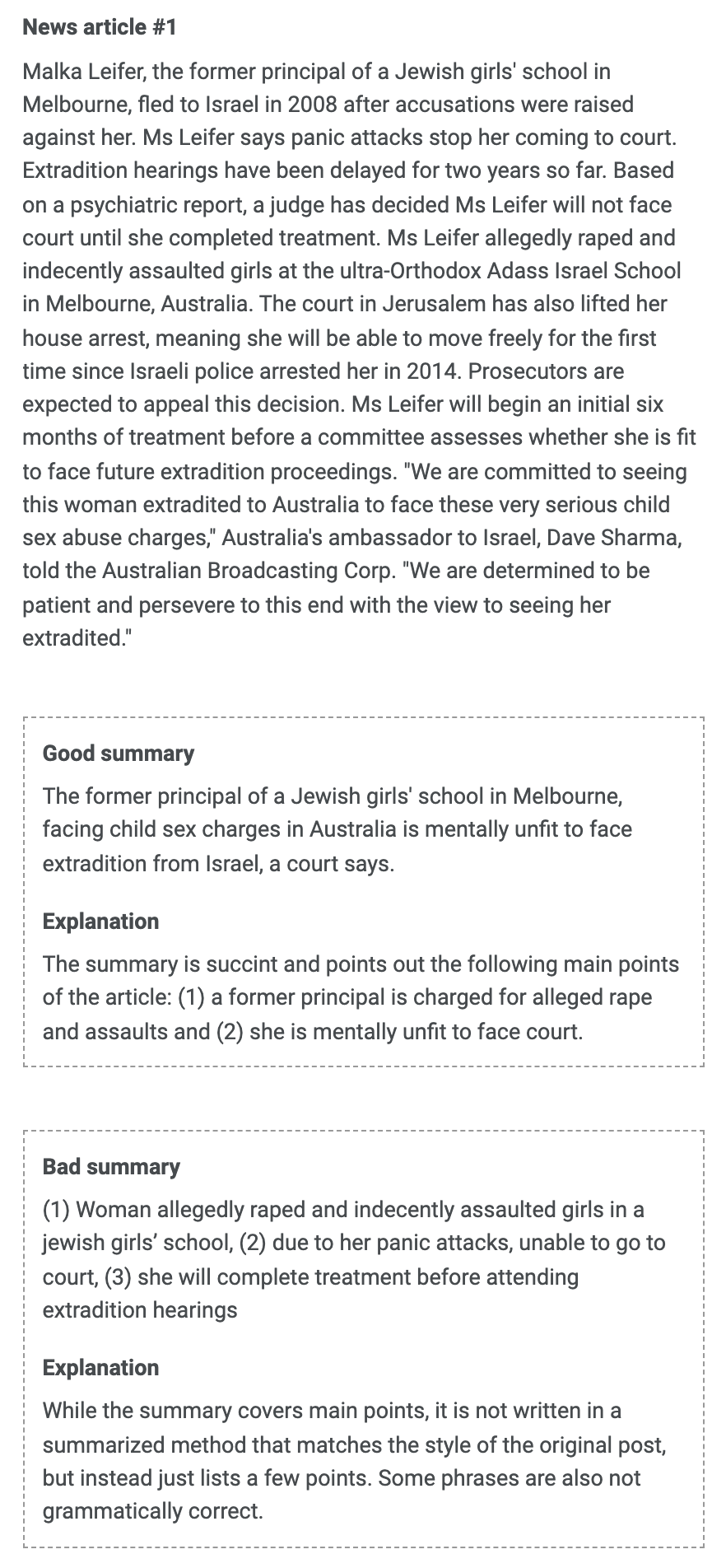}
    \end{subfigure}
    \caption{\dataxsum tutorial first example with good and bad explanations.}
    \label{fig:tutorial_example2}
\end{figure}

\begin{figure}
    \centering
    \begin{subfigure}{0.5\textwidth}
        \centering
        \includegraphics[width=\textwidth]{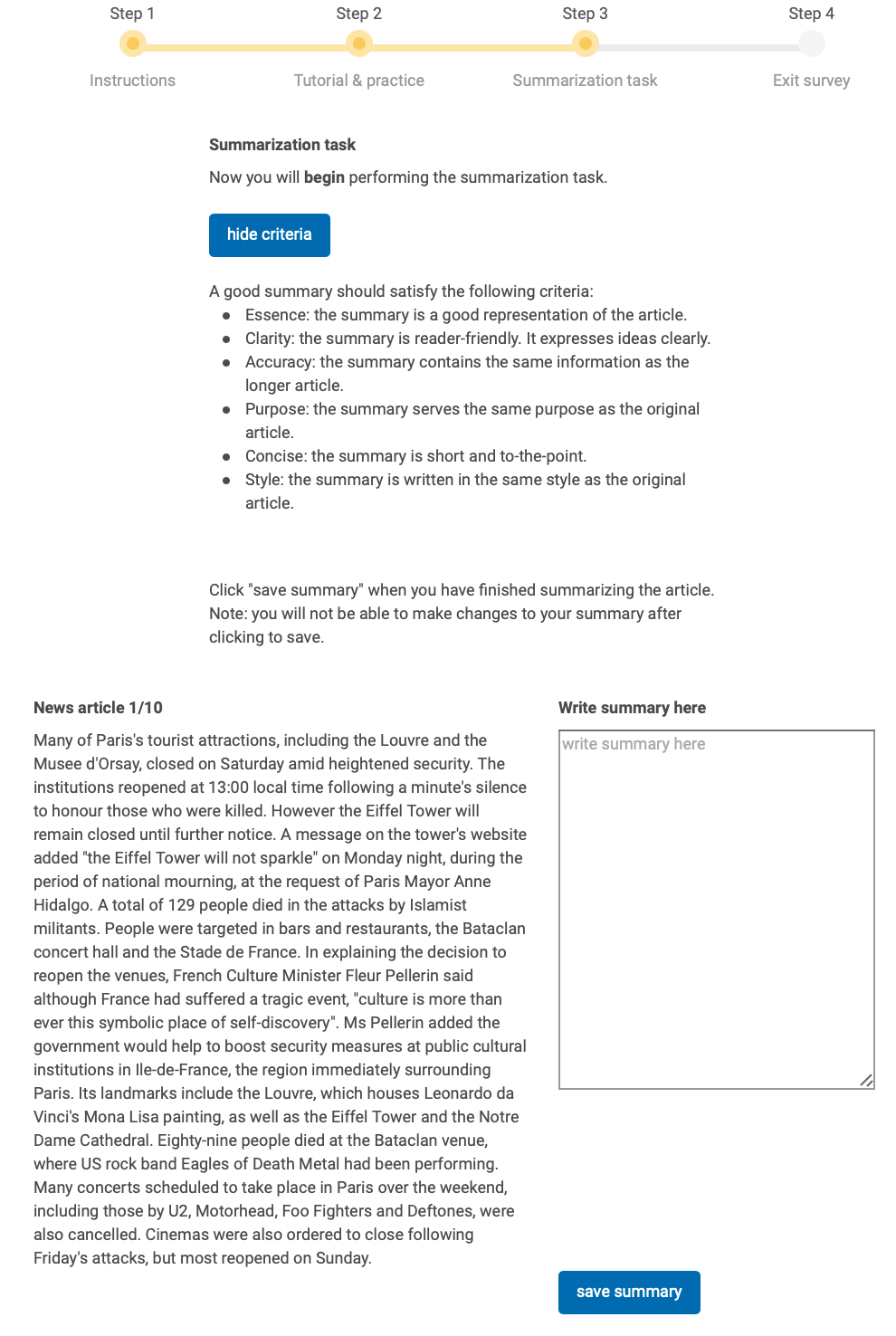}
        \caption{Task interface.}
        \label{fig:task_interface}
    \end{subfigure}
    \begin{subfigure}{0.5\textwidth}
        \centering
        \includegraphics[width=\textwidth]{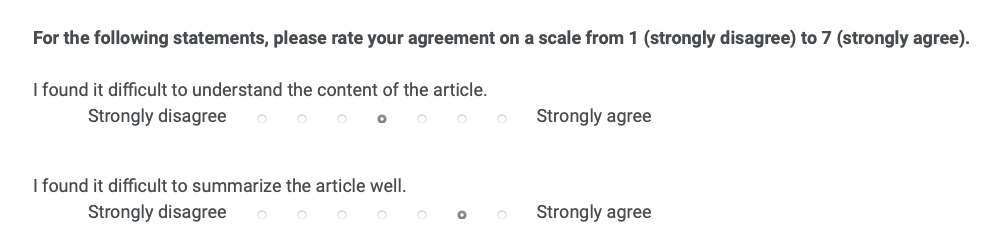}
        \caption{Questions regarding the summarization task.}
        \label{fig:task_questions}
    \end{subfigure}
    \caption{Task interface and questions on the summarization task.}
    \label{fig:task_interface_questions}
\end{figure}

\subsubsection{Summarization Study Procedure}
\label{app:phase1_procedure}
During the study, participants completed three phases: (1) instructions, tutorial, and practice; (2) summarization task; and (3) post-task survey.
During the \textit{instructions, tutorial, and practice} phase, participants reviewed task instructions, the criteria for writing a good summary (\tabref{tb:summary_criteria}), and examples of good and bad summaries with explanations (\figref{fig:tutorial_example1} and \figref{fig:tutorial_example2}).
Participants then applied what they learned and practiced to write a good summary.
To refrain from any confusion, the interface in the practice phase is exactly the same as the actual summarization task phase.
Participants have access to the criteria as guidance at any time during the task (\figref{fig:task_interface}).
%
Participants then performed the summarization task as described in \secref{sec:procedure}.
After completing each summary, participants rated their agreement for the following statements (\figref{fig:task_questions}): (1) I found it difficult to understand the content of the document; (2) I found it difficult to summarize the document well.
%
Finally, participants responded to an exit survey before ending the study, where they answered questions regarding task ease, frustration, their familiarity with the domain (i.e., Reddit, British news and culture), if the provided assistance was useful, and what they liked and disliked about the summarization task and interface.


\subsection{Human Evaluation of the Collected Summaries}
\label{app:eval}

We used human evaluation to assess the quality of the summaries we collected during our study.
Each annotation task involved reading a news article or Reddit post and evaluating six different summaries for that document.

\subsubsection{Example Summaries}
\label{app:example_summary_type}

We evaluated six types of summaries: 
\begin{enumerate}
    \item \textbf{Manual}. It was written without assistance.
    \item \textbf{AI-generated}. The provided summary was generated by the Pegasus model.
    \item \textbf{Human reference}. The provided summary is from the original dataset.
    \item \textbf{AI + post-edit}. It was written by a human who was shown a AI-generated summary.
    \item \textbf{Human + post-edit.} It was written by a human who was shown a reference summary.
    \item \textbf{Random.} It was generated by randomly selecting two sentences from another dataset. This summary helped to weed out annotators who did not pay attention. 
     For instance, if the participant sees a Reddit post, the \textit{Random} summary is a summary generated from a news article from the \dataxsum dataset.
\end{enumerate}

\tabref{tb:example_summaries} demonstrate the six example summaries for one document.

\begin{table*}[h!]
    \small
    \centering
    \begin{tabular}{p{0.15\textwidth}p{0.4\textwidth}p{0.4\textwidth}}
    \toprule
    Condition & \datareddit & \dataxsum \\ \midrule
    
    Document & so today i went to the zoomarine in algarve, portugal and my family decided to go watch the interactive pirate show (basicly pirates play around with the crowd), so far so good... we sit down and to my right, this long hair, tatooed, bearded, baggy clothes, pirate looking guy sits down by himself. a bit after my mom without moving her head asks me if i have seen the pirate "hiding" which i quietly answer to with "yeah". here i am thinking "nice! he is probably gonna let me take a photo with him!". so, i get my camera and ask him if i could take a picture with him. well, just when i am starting to pose for the picture, he stands up and starts cursing giberish in spannish (in other words: not in portuguese) and got everyone in the audience and my entire family looking at me and him like it is part of the show... at this point my mom looks even more confused than me thinking what i could have done. needless to say, later in the show the pirate my mom was talking about was 2 rows down from us. & The former Tottenham player, 29, is the Spanish club\'s first signing since Neymar left to join Paris St-Germain in a world record transfer. Meanwhile, Barca\'s Uruguayan forward Luis Suarez will be out for "four to five weeks" after he was injured in the Super Cup defeat by Real Madrid. He will also miss World Cup qualifiers against Argentina and Paraguay. Paulinho joined Tottenham for £17m from Corinthians in 2013, before moving to China in 2015. He helped Evergrande win last season\'s Chinese Super League and leaves the team top of the table in the current campaign. Paulinho said: "You have to face challenges with courage. I will try to do my job and I am prepared. It\'s a very satisfying moment. The dream I have been looking for has come true. I will give everything." Barcelona are also keen on Liverpool midfielder Philippe Coutinho and Borussia Dortmund forward Ousmane Dembele. \\ \midrule
    
    Manual & My family and I went to an interactive pirate show and accidentally mistook a shabbily dressed audience member for a secret pirate actor by asking to take a picture with him. & This week in sport's news, Spanish club has signed its first player since Neymar left to join Paris St-Germain, he gives motivation words ahead of season. While, Uruguayan forward Luis Suarez will be out recovering from an injury from Super Cup defeat. \\ \midrule
    
    AI post-edit & asked a pirate if i could take a picture with him, he started cursing in spannish and got everyone in the audience and my entire family looking at me and him like it is part of the show. & Barcelona have signed Brazil midfielder Paulinho from Chinese club Guangzhou Evergrande for an undisclosed fee. \\ \midrule
    
    Human post-edit & went to pirate show, saw a pirate looking guy, tried to take selfie with him, it was an evil hipster. & Barcelona have signed Brazil midfielder Paulinho from Chinese club Guangzhou Evergrande for 40m euro (£36.4m). \\ \midrule
    
    AI-generated & today at the zoomarine in algarve, portugal i wanted to take a picture with a pirate but he started cursing and then things got weird & Barcelona have signed Brazil midfielder Paulinho from Chinese club Guangzhou Evergrande for an undisclosed fee. Suarez to be out four to five weeks with an injury. \\ \midrule
    
    Human reference & Was watching an interactive pirate show and thought the guy next to me was an actor. Asked to take a selfie and got yelled at in Spanish. He wasn't an actor. & Barcelona signs Paulinho, while also seeking out Philippe Coutinho and Ousmane Dembele. Suarez is expected to miss two World Cup qualifying games due to injury. \\ \midrule
    
    Random & Avon and Somerset Police have named the victim as Matthew Symonds, 34, of no fixed address in Swindon, and said his death was being treated as unexplained. A post-mortem examination is due to be carried out later. & would anyone really mind if i just kept internet explorer (**my school computers have no other internet browsers help**) open on the side and just read some funny tifus??? the whole class bursts out laughing.\\
    \bottomrule
    \end{tabular}
    \caption{\datareddit and \dataxsum example summaries.
    }
    \label{tb:example_summaries}
\end{table*}

\begin{figure}[h!]
    \centering
    \begin{subfigure}{0.5\textwidth}
        \centering
        \includegraphics[width=\textwidth]{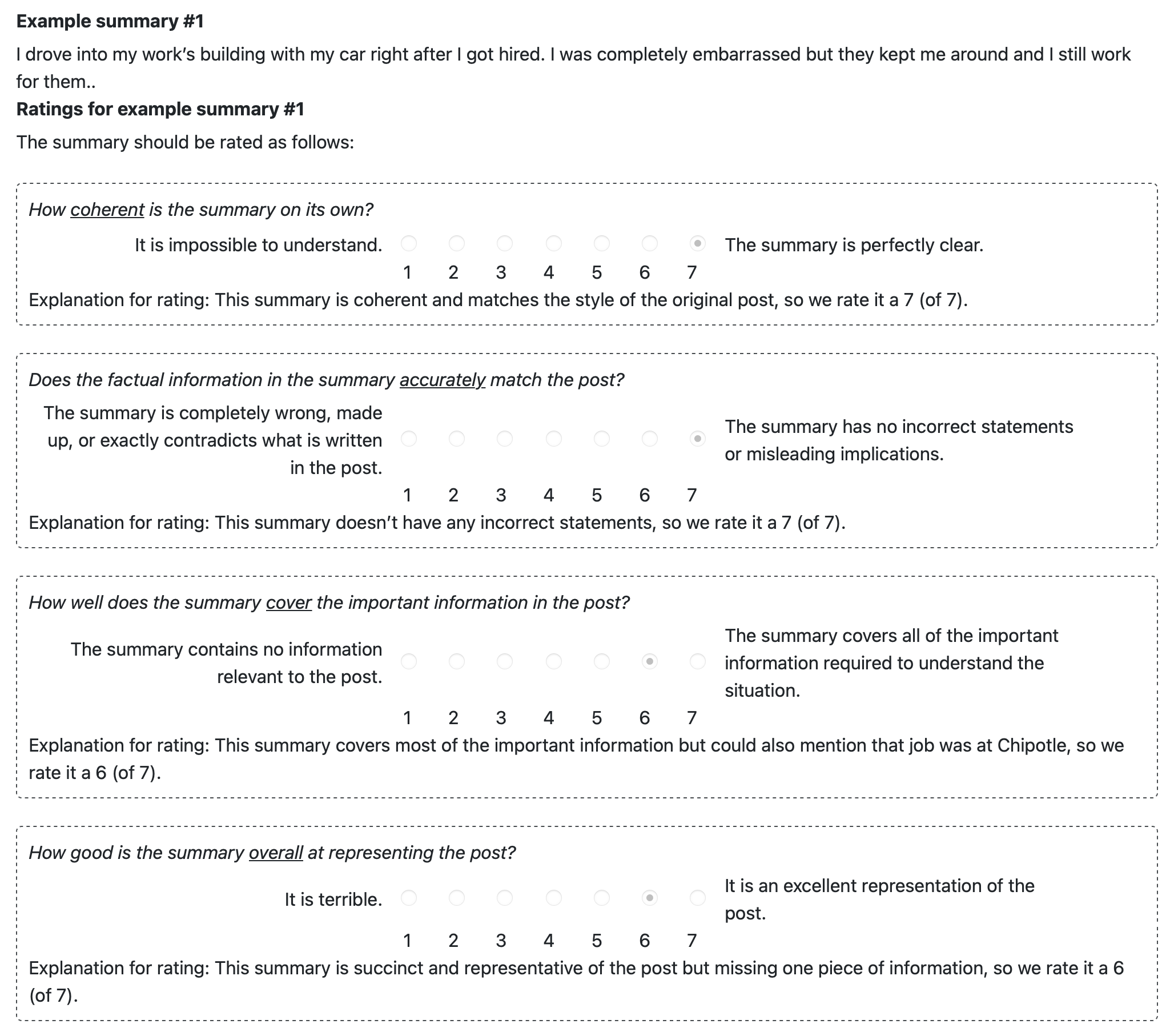}
        \caption{Tutorial interface.}
        \label{fig:evaluation_interface_tutorial}
    \end{subfigure}
    \begin{subfigure}{0.5\textwidth}
        \centering
        \includegraphics[width=\textwidth]{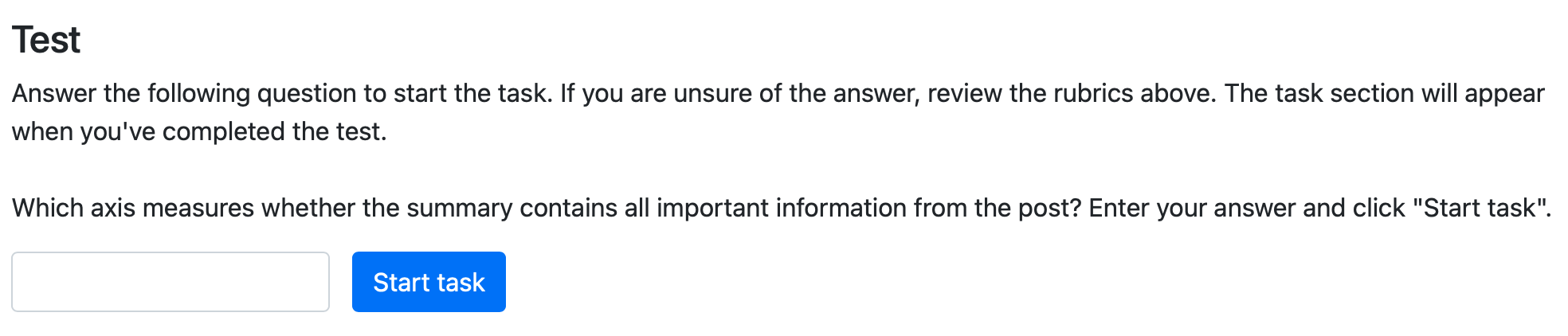}
        \caption{Attention check interface.}
        \label{fig:evaluation_interface_attention_check}
    \end{subfigure}
    \begin{subfigure}{0.5\textwidth}
        \centering
        \includegraphics[width=\textwidth]{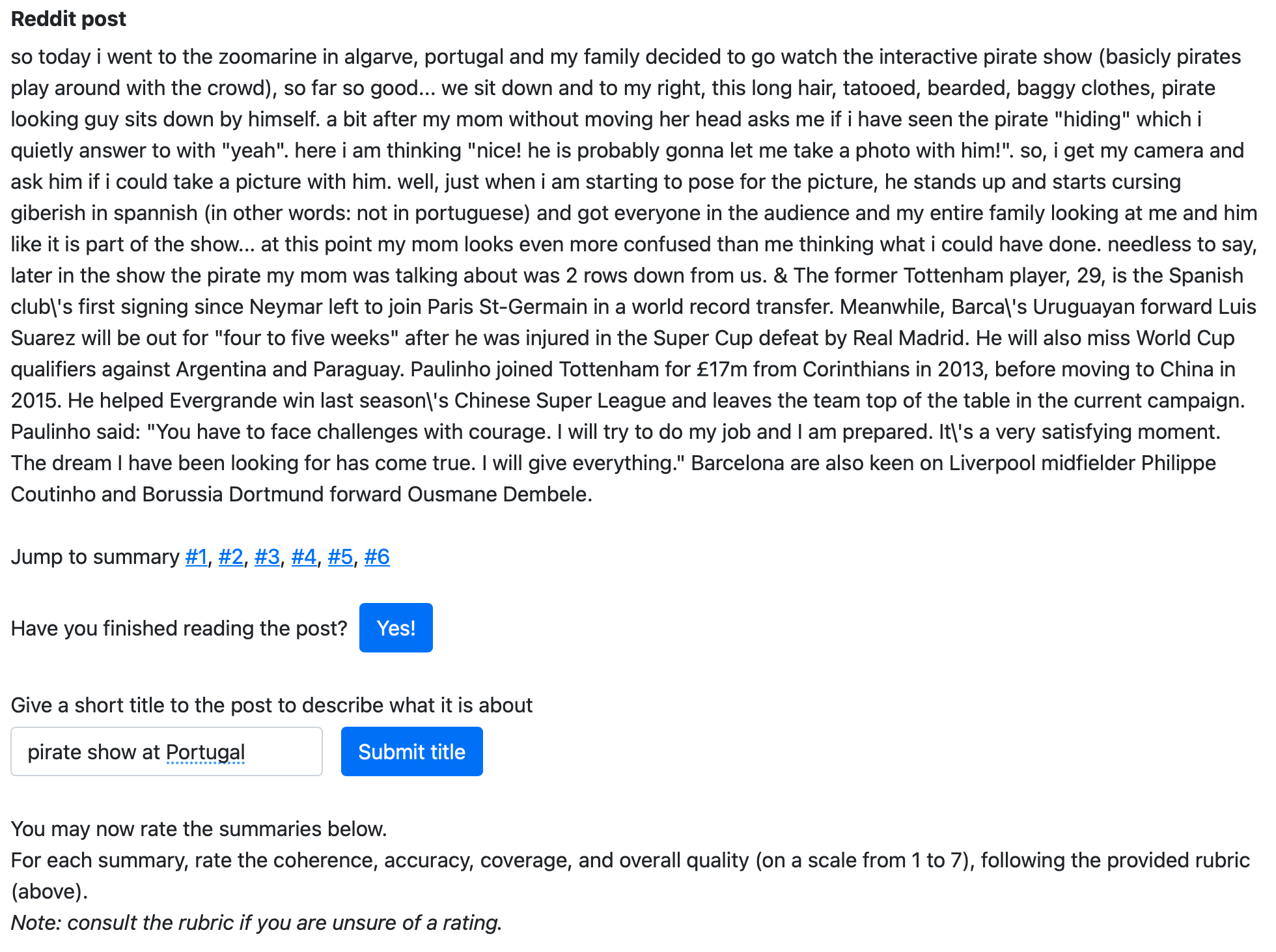}
        \caption{Instructions are given before the actual task.}
        \label{fig:evaluation_interface_evaluate_part1}
    \end{subfigure}
    \begin{subfigure}{0.5\textwidth}
        \centering
        \includegraphics[width=\textwidth]{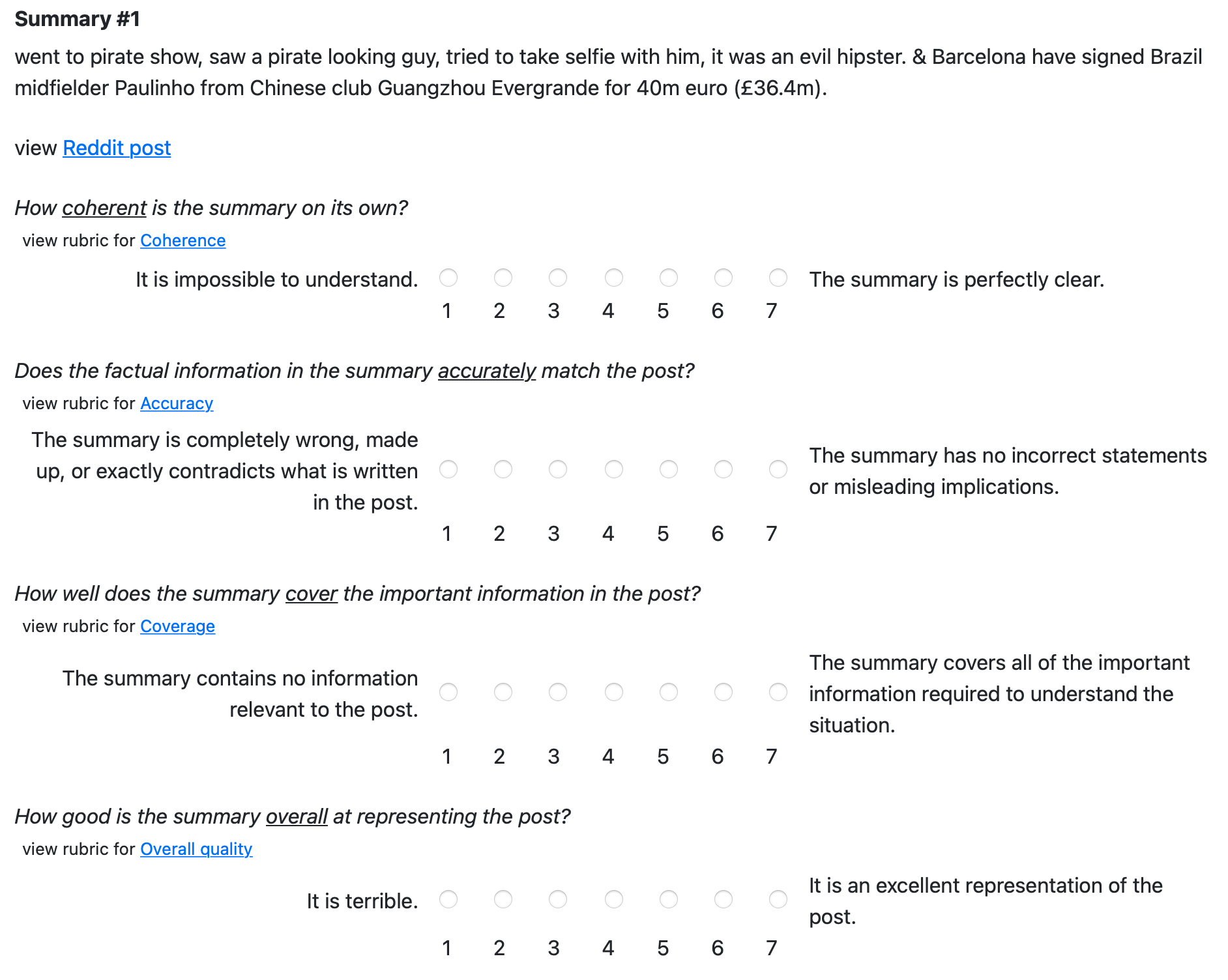}
        \caption{Evaluation interface.}
        \label{fig:evaluation_interface_evaluate_part2}
    \end{subfigure}
    \caption{Annotation interfaces.}
    \label{fig:evaluation_interface}
\end{figure}

\subsubsection{Summary Evaluation Procedure}
\label{app:evaluate_summary}
The annotator went through two phases during the human evaluation task: (1) tutorial (\figref{fig:evaluation_interface_tutorial}) and attention check (\figref{fig:evaluation_interface_attention_check}); and (2) evaluating summaries (\figref{fig:evaluation_interface_evaluate_part1} and \figref{fig:evaluation_interface_evaluate_part2}).
A tutorial with two examples was provided at the start of the task to teach the participant how to evaluate a summary.
To further solidify their understanding, we also included two examples with ratings and explanations.
Explanations were curated by the researchers and iterated a few times upon getting feedback.
We included an attention check open-ended question to ensure that the participant read through and understood the tutorial.
The task would only begin when they give the correct answer.
Before the actual task, to ensure that annotators read the document, we asked them to write a short title after they read it.
During the task, anchors for the original post and definition of axis are easily accessible, allowing annotators to refer to them whenever they wanted to.
A HIT \footnote{\href{https://docs.aws.amazon.com/AWSMechTurk/latest/RequesterUI/mechanical-turk-concepts.html}{Amazon Mechanical Turk Concepts}} included reading a document and evaluating six different types of summaries of it.
For each HIT an annotator does, they are paid 
\$1.50.
After removing annotators that failed the attention check, the average time taken to complete a HIT is 13.5 minutes (SD=5.7).
Although the hourly rate may seem low,
we learned from the annotators that turkers tend to open up to 25 tabs of HIT while working.
As such, the time taken also included idle time, meaning that the actual average time taken could be less than 13.5 minutes.
Additionally, annotators did 9.4 HITs on average.


\subsubsection{Eliminating the bad apples.}
\label{app:quality_control}

To ensure high-quality evaluations, we performed quality control to eliminate annotators with validation procedures: attention check with a ``random'' summary, batched deployment, and removing outliers. We detail each of these procedures in the following.

\para{Attention check: \textit{random} summary.}
We incorporated an attention check to weed out annotators who did not pay attention and simply clicked through the rating scales.
For this, we inserted a \textit{random} summary to rate, which was generated by randomly selecting two sentences from the opposite dataset.
Further, we randomize the order of summaries, ensuring that the \textit{random} summary would not always appear in the same position.
Since the \textit{random} summary is created with sentences from a document from an entirely different dataset, it is fair to assume that the content will not cover nor be accurate as a summary. 
Therefore, we eliminated annotators (and discarded their responses) who did not give a rating of 1 to both \textit{coverage} and \textit{accuracy}.

\para{Batched deployments.}
To ensure only high-performing annotators participated in our evaluations and maintain the integrity of our results, we followed a batched deployment procedure opposed to deploying all evaluations at once.
We deployed a total of 1200 evaluations (i.e., 120 HITs x 2 datasets x 5 samples) on Amazon Mechanical Turk, splitting the assignments into 10 batches.\
At the end of a batch deployment, annotators who failed the attention check (\textit{random} summary)
had their qualification revoked
and not allowed to accept future HITs for our evaluation task. 
A total of 113 annotators completed the 1200 evaluations.

\para{Removing outliers.}
\label{app:remove_outliers}
Finally, while we had 5 annotators evaluating a summary, not all five ratings were taken into consideration.
We removed any outlier ratings following a standard approach: 1.5 more or less than the inter quartile range (IQR).
\tabref{tb:ratings_variance} show the variance for each condition per dataset after removing outliers.

\subsubsection{Thematic Coding}
\label{app:thematic_coding}
We performed thematic coding to analyze the open-ended responses in our study.
A researcher in the team first manually coded the data and iteratively developed themes from answers of each question. 
For example, "lack of context knowledge" is one of the themes developed from the answers to the question on why participants rated the task as difficult. 
The researcher then discussed the themes with the team, merged and updated the themes, then re-coded the data again. 
To validate the coding results, a second researcher also coded the data based on the themes developed by the first researcher. 

\subsubsection{Summary Quality Criteria: Coherence, Accuracy, Coverage, and Overall}
Annotators were tasked to evaluate a summary according to four axes on a scale of 1 to 7 \citep{stiennon2020learning}.
The definition of each axis is listed as below:
\begin{enumerate}
    \item \textbf{Coherence}. A summary is \textit{coherent} if, when read by itself, it's easy to understand and free of English errors. A summary is not coherent if it's difficult to understand what the summary is trying to say. Generally, it's more important that the summary is understandable than it being free of grammar errors. 
    \item \textbf{Accuracy}. A summary is \textit{accurate} if it doesn't say things that aren't in the article, it doesn't mix up people, and generally is not misleading. If the summary says anything at all that is not mentioned in the article or contradicts something in the article, it should be given a maximum score of 5.
    \item \textbf{Coverage}. A summary has good \textit{coverage} if it mentions the main information from the article that's important to understand the situation described in the article. A summary has poor coverage if someone reading only the summary would be missing several important pieces of information about the situation in the article. A summary with good coverage should also match the purpose of the original article (e.g. to ask for advice). 
    \item \textbf{Overall}. This can encompass all of the above axes of quality, as well as others you feel are important. If it’s hard to find ways to make the summary better, give the summary a high score. If there are lots of different ways the summary can be made better, give the summary a low score. 
\end{enumerate}

\begin{table}[h!]
    \small
    \centering
    \begin{tabular}{p{0.1\textwidth}p{0.15\textwidth}p{0.15\textwidth}}
    \toprule
     & Condition & Rating (\_sigma)\\ \midrule
    \datareddit & Manual & 0.81 \\
     & AI post-edit & 0.95 \\
     & Human post-edit & 1.05\\
     & AI-generated & 0.63 \\
     & Human reference & 0.66 \\ \midrule
    \dataxsum & Manual & 0.85 \\
     & AI post-edit & 0.84 \\
     & Human post-edit & 0.89 \\
     & AI-generated & 1.02 \\
     & Human reference & 0.63 \\
    \bottomrule
    \end{tabular}
    \caption{Variance between ratings for each condition per dataset for the Overall rating.}
    \label{tb:ratings_variance}
\end{table}

\subsubsection{Research Experiment Ethics}
\label{app:ethics}
Participants from Upwork and annotators from Amazon Mechanical Turk were aware of how the data collected would be used.
They were assured that no personally identifiable information was collected from them.
For participants on Upwork, the written summaries and exit survey responses were collected from them.
Similarly, for annotators on Amazon Mechanical Turk, only responses and ratings were collected.
Before working on the task, participants and annotators were made to read a description of the task and working on the task meant that they were aware of what was collected.

\subsection{Results}
\label{app:results}

\subsubsection{Ratings on Coherence, Accuracy, and Coverage}
\label{app:acc_cov_ratings}
\begin{figure}[h!]
    \centering
    \begin{subfigure}{0.5\textwidth}
        \centering
        \includegraphics[width=\textwidth]{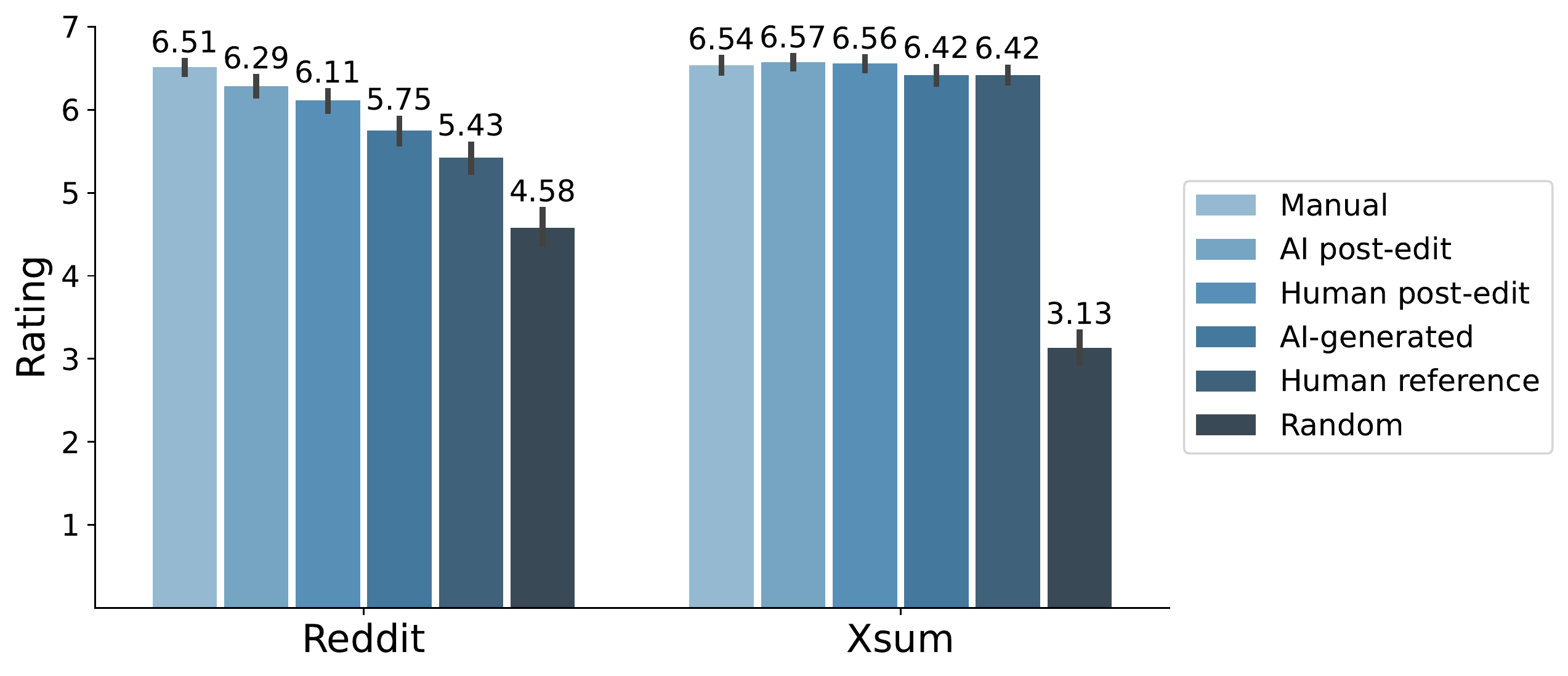}
        \caption{Coherence rating.}
        \label{fig:quality_coherence_combined}
    \end{subfigure}
    \begin{subfigure}{0.5\textwidth}
        \centering
        \includegraphics[width=\textwidth]{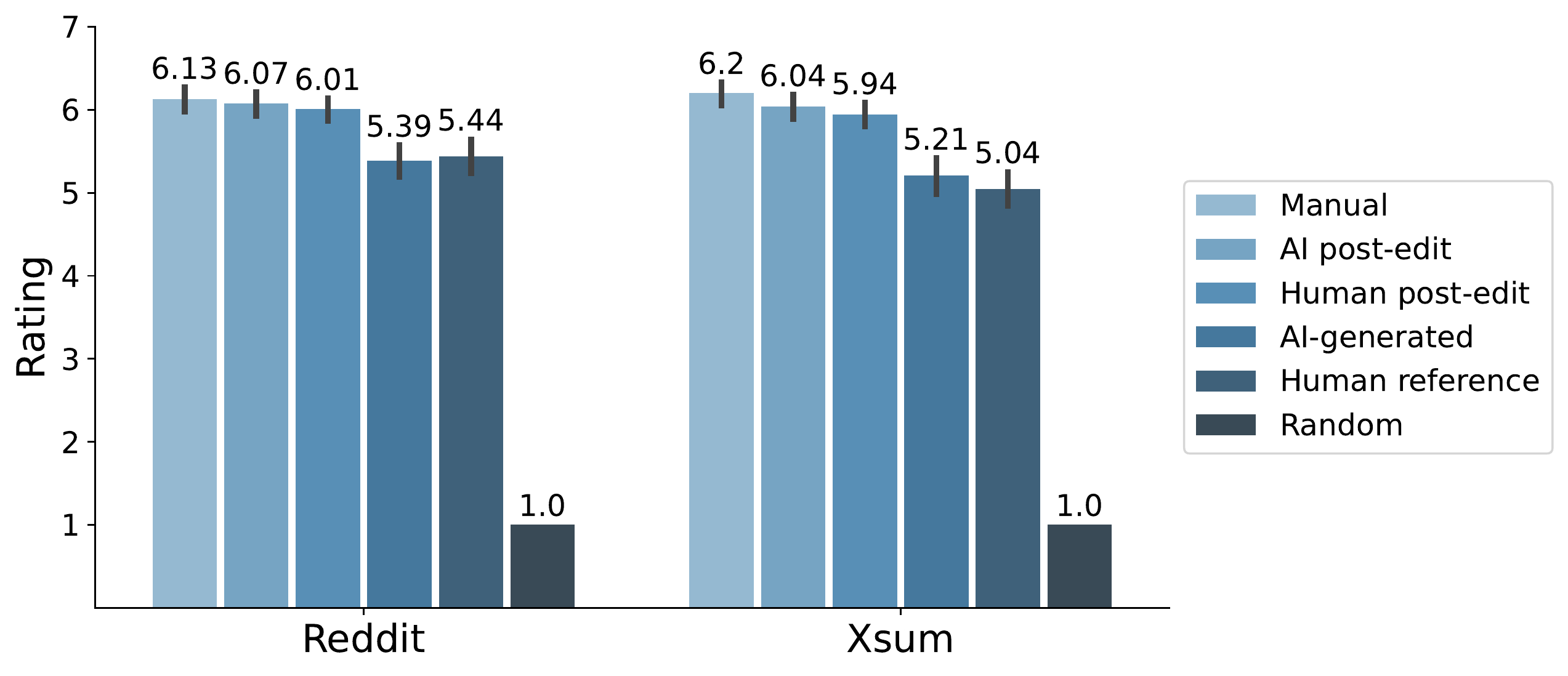}
        \caption{Accuracy rating.}
        \label{fig:quality_accuracy_combined}
    \end{subfigure}
    \begin{subfigure}{0.5\textwidth}
        \centering
        \includegraphics[width=\textwidth]{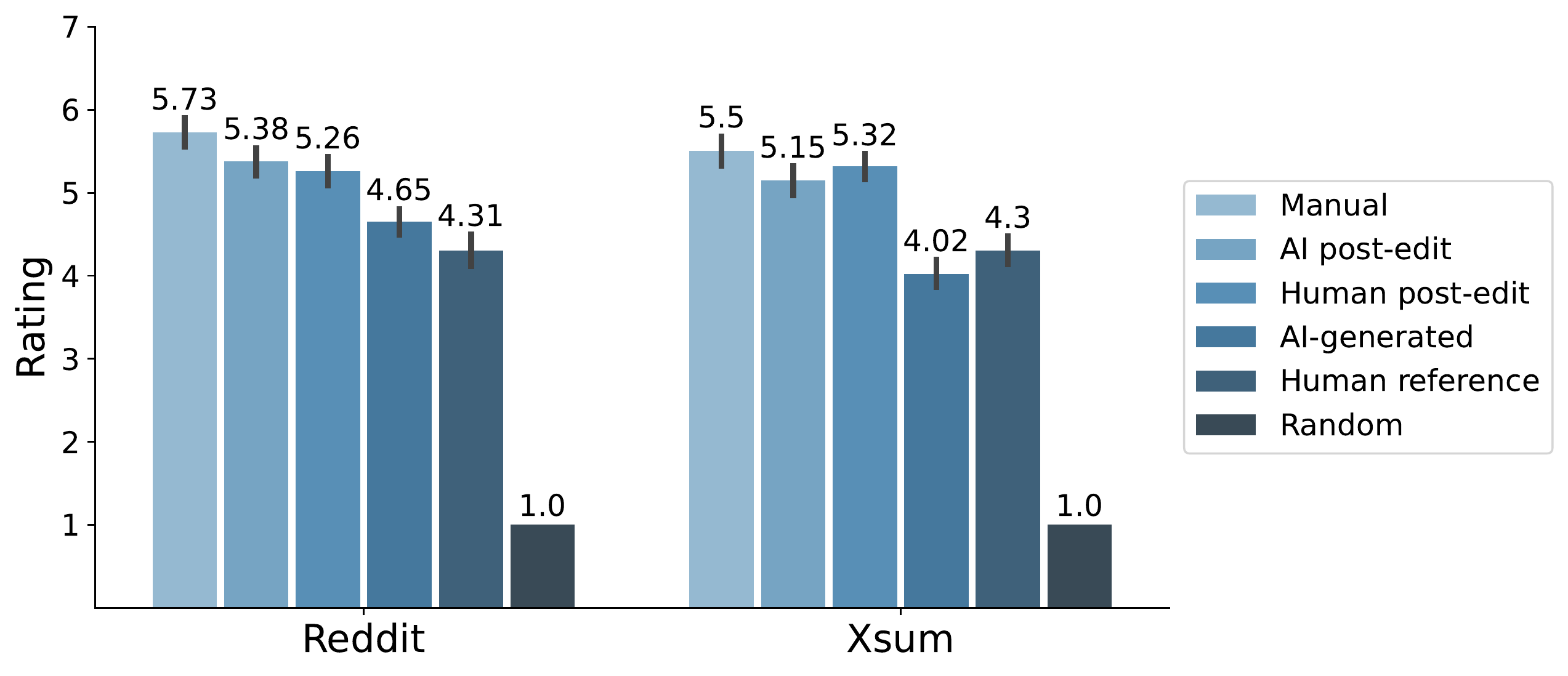}
        \caption{Coverage rating.}
        \label{fig:quality_coverage_combined}
    \end{subfigure}
    \caption{Human evaluation rating on \textit{coherence}, \textit{accuracy}, and \textit{coverage}. 
    }
    \label{fig:quality_accuracy_coverage}
\end{figure}

In the main paper, we reported only \textit{overall} ratings.
\figref{fig:quality_accuracy_coverage} shows the plots for \textit{coherence}, \textit{accuracy}, and \textit{coverage} ratings.
Both \datareddit and \dataxsum summaries produced by participants in the \condmanual condition were rated highest \textit{accuracy} and \textit{coverage} quality.

In \datareddit, \textit{AI assistance} significantly impacted coherence ($p<.01, F=15.4$) and coverage ($p<.01, F=7.9$) but not accuracy ($p=.48, F=.7$): \condmanual summaries outperformed summaries produced by participants in both the \condai ($p=.01$ for coherence, $p=.01$ for accuracy) and \condhuman ($p<.01$ for coherence, $p=.01$ for accuracy) conditions.

In \dataxsum, \textit{AI assistance} significantly impacted accuracy ($p=.03, F=3.5$) and coverage ($p=.02, F=3.8$) but not coherence ($p=.84, F=.17$): \condmanual summaries outperformed summaries produced by participants in \condhuman ($p=.03$ for accuracy) and \condai ($p=.01$ for coverage).


 \subsubsection{Correlation between Edit Distance and Summary Quality}
 \label{app:edit_distance}
\begin{figure}
    \centering
    \begin{subfigure}{0.45\textwidth}
        \centering
        \includegraphics[width=\textwidth]{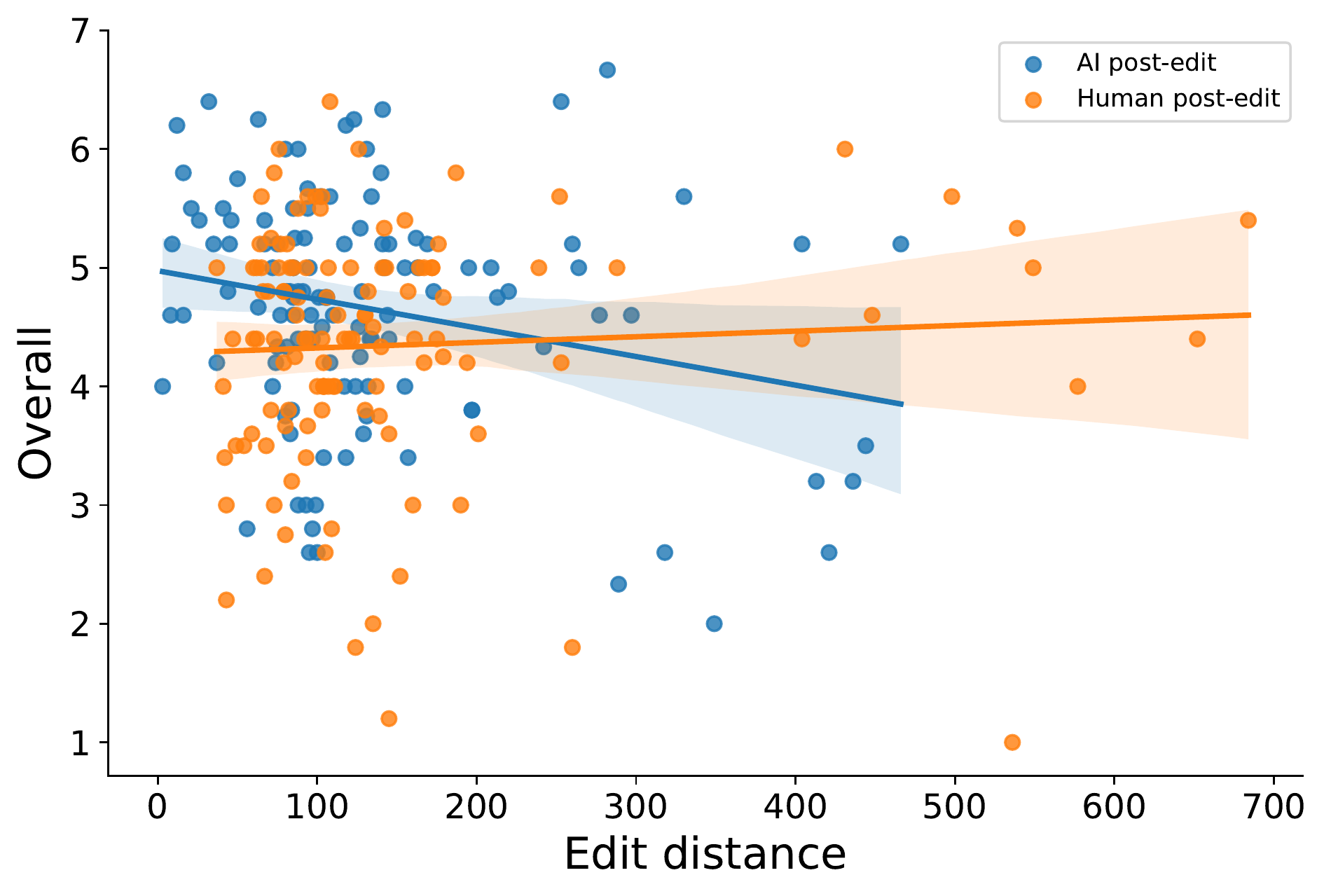}
        \caption{\datareddit}
        \label{fig:edit_dist_overall_reddit}
    \end{subfigure}
    \begin{subfigure}{0.45\textwidth}
        \centering
        \includegraphics[width=\textwidth]{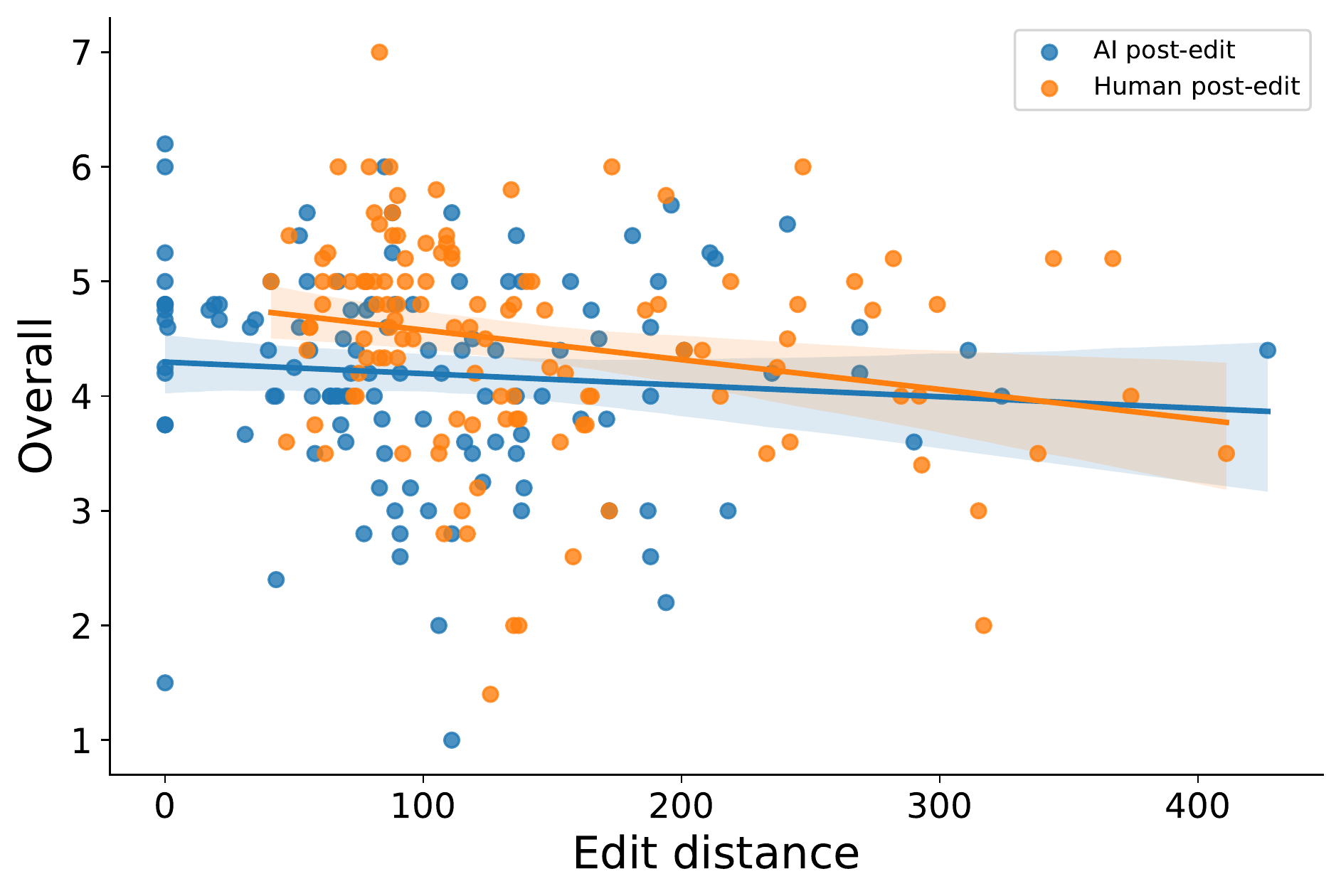}
        \caption{\dataxsum}
        \label{fig:edit_dist_overall_xsum}
    \end{subfigure}
    \caption{Edit distance vs. Overall rating}
    \label{fig:edit_dist_overall}
\end{figure}

We compared edit distance and overall summary rating and found weak to no correlation ($p = 0.4, \rho = -0.05$ 
 for \datareddit and $p=.02, \rho = -0.2$ for \dataxsum) between the two factors. 
 For \dataxsum, while this suggests that the bigger the edit distance, the poorer the overall summary rating, the correlation score is very small.
 On the other hand, there is no correlation between edit distance and overall summary rating in \datareddit.
 \figref{fig:edit_dist_overall} shows the plots for both datasets.

\subsubsection{User Experience}
In the main paper, we reported insights from qualitative analysis for task difficulty, frustration, and assistance utility (\secref{sec:user_experience}).
We also conducted thematic coding on why participants enjoyed working on the summarization task.

\para{The summarization task was enjoyable and educational.}
Many participants enjoyed working on the task, describing the experience as educational (e.g., P14 (\dataxsum, \condmanual), ``it made me think about the information I had read and how to best condense it''). Others enjoyed reading the original text (e.g., P53 (\datareddit, \condai), ``these stories are quite interesting, the summaries make me make sure I understood what I just read"), and felt a sense of achievement when finished (e.g., P36 (\dataxsum, \condhuman), ``it was satisfying to reduce a block of text down to a succinct sentence or two''). 

\end{document}